\pdfoutput=1
\documentclass[11pt]{article}
\usepackage[table]{xcolor}
\definecolor{verylightgray}{gray}{0.82}
\usepackage[final]{acl}

\usepackage{times}
\usepackage{latexsym}
\usepackage{multirow}
\usepackage{adjustbox}
\usepackage{booktabs} % Include this in your preamble
\usepackage{subcaption}
\usepackage[T1]{fontenc}
\usepackage[utf8]{inputenc}

\usepackage{microtype}

\usepackage{inconsolata}

\usepackage{graphicx}

\title{\textit{\MakeUppercase{D\scalebox{.75}{ECOR}}}: Improving Coherence in L2 English Writing with a Novel Benchmark for Incoherence Detection, Reasoning, and Rewriting}

\author{
 \textbf{Xuanming Zhang\textsuperscript{1}},
 \textbf{Anthony Diaz\textsuperscript{2}},
 \textbf{Zixun Chen\textsuperscript{1}}, \\
 \textbf{Qingyang Wu\textsuperscript{1}},
 \textbf{Kun Qian\textsuperscript{1}},
 \textbf{Erik Voss\textsuperscript{1}},
 \textbf{Zhou Yu\textsuperscript{1}}
\\
\\
 \textsuperscript{1}Columbia University,
 \textsuperscript{2}University of California, Davis
\\
\small{
 \texttt{
   \{xz2995, zc2738, zy2461\}@columbia.edu, antdiaz@ucdavis.edu
 }}
}

\begin{document}
\maketitle
\begin{abstract}

% In NLP, the assessment of coherence is primarily conducted on machine-generated texts, evaluating their logical consistency and meaningfulness to the readers. 

% However, little effort has been made to assess coherence in human-written texts, an endeavor that is highly beneficial for language learners aiming to enhance their writing skills. 

Coherence in writing, an aspect that second-language (L2) English learners often struggle with, is crucial in assessing L2 English writing.
Existing automated writing evaluation systems primarily use basic surface linguistic features to detect coherence in writing. However, little effort has been made to correct the detected incoherence, which could significantly benefit L2 language learners seeking to improve their writing. To bridge this gap, we introduce \textit{DECOR}, a novel benchmark that includes expert annotations for detecting incoherence in L2 English writing, identifying the underlying reasons, and rewriting the incoherent sentences. To our knowledge, \textit{DECOR} is the first coherence assessment dataset specifically designed for improving L2 English writing, featuring pairs of original incoherent sentences alongside their expert-rewritten counterparts. Additionally, we fine-tuned models to automatically detect and rewrite incoherence in student essays. We find that incorporating specific reasons for incoherence during fine-tuning consistently improves the quality of the rewrites, achieving a result that is favored in both automatic and human evaluations.\footnote{Data and code available: \url{https://github.com/BillyZhang24kobe/writing2coherence}}

\end{abstract}

\section{Introduction}

% ai tutoring is popular among language learners; as a supplement of expensive human instructor feedback, among these tools, there are GEC tools

% there are AI tools to support language learners; these tools include A B C. no discourse level 

% With the proliferation of computer-assisted English learning tools, learners have the ability to be more autonomous while simultaneously not having to rely on careful feedback from a language instructor in order to improve their second-language (L2) writing skills. 
Automatic English writing tools have gained extensive popularity among second-language (L2) learners. These tools serve as a cost-effective supplement to traditional, expensive human tutoring, providing learners with timely and constructive feedback. Much progress in this area includes automatic grammar correction systems (\citealt{omelianchuk2020gector,yasunaga2021lm,tarnavskyi2022ensembling,cao2023mitigating}) and tools to improve the vocabulary usage of learners (\citealt{johnson2016vocabulary,gonzalez2017contribution,zhang2024prolex}). However, these tools primarily focus on the word and sentence-level issues that affect L2 writing rather than discourse-level issues.

An aspect of L2 writing that could also benefit from automated tools is the overall textual coherence which is a requirement to efficiently convey one’s ideas. To improve L2 writing skills, whether it is part of a course assessment or standardized test of English ability, learners are often required to carefully organize their thoughts in response to a predetermined writing prompt. Previous research has identified coherence as a crucial feature to measure when assessing L2 writing proficiency, as it is an aspect that students often struggle with (\citealt{schneider1990analyzing,bitchener2006perceptions,cooley1995writing,lorenz1999learning}). Current automated writing evaluation tools primarily provide learners with scores that indicate the level of coherence in their writing \cite{naismith2023automated}. They primarily detect coherence with simple surface linguistic features, such as syntax and parts of speech \cite{mcnamara2010coh,crossley2016tool}. 
However, merely detecting coherence in writing is insufficient to help L2 English writers enhance their writing. An automated system capable of detecting incoherence in L2 writing, identifying the underlying reasons, and correcting the incoherent sentences would be immensely valuable for both language learners and instructors. However, the absence of a benchmark dataset specifically designed for incoherence detection, reasoning, and rewriting in L2 English essays significantly impedes the development of such systems.

% An automated system that can not only detect incoherence in L2 writing but also suggest the reasons behind the incoherence and correct the incoherent sentences would be tremendously beneficial for both language learners and instructors. Nevertheless, there is no existing benchmark dataset that is tailored for incoherence detection, reasoning, and rewriting in L2 English essays, thus hampering the development of such automated systems.

% To facilitate this effort, there is a need for a dataset specifically designed to detect, reason, and rewrite incoherence in learner-written essays.

% to the best of our knowledge, there is no existing benchmark dataset that is tailored for assessing coherence in L2 English writing, hampering the development of automated systems that could detect and rewrite incoherence in learner-written essays. 

% We specifically recruited two expert annotators who are both associate professors with extensive experience in teaching English as a foreign language and have advanced degrees in Applied Linguistics.

% perform the following three tasks: 1) detect if the current sentence $S$ is incoherent with the context $C$, 2) identify specific reasons that cause the incoherence, and 3) rewrite the incoherent sentences based on the identified reasons.

\begin{figure}[t]
  \includegraphics[width=\columnwidth]{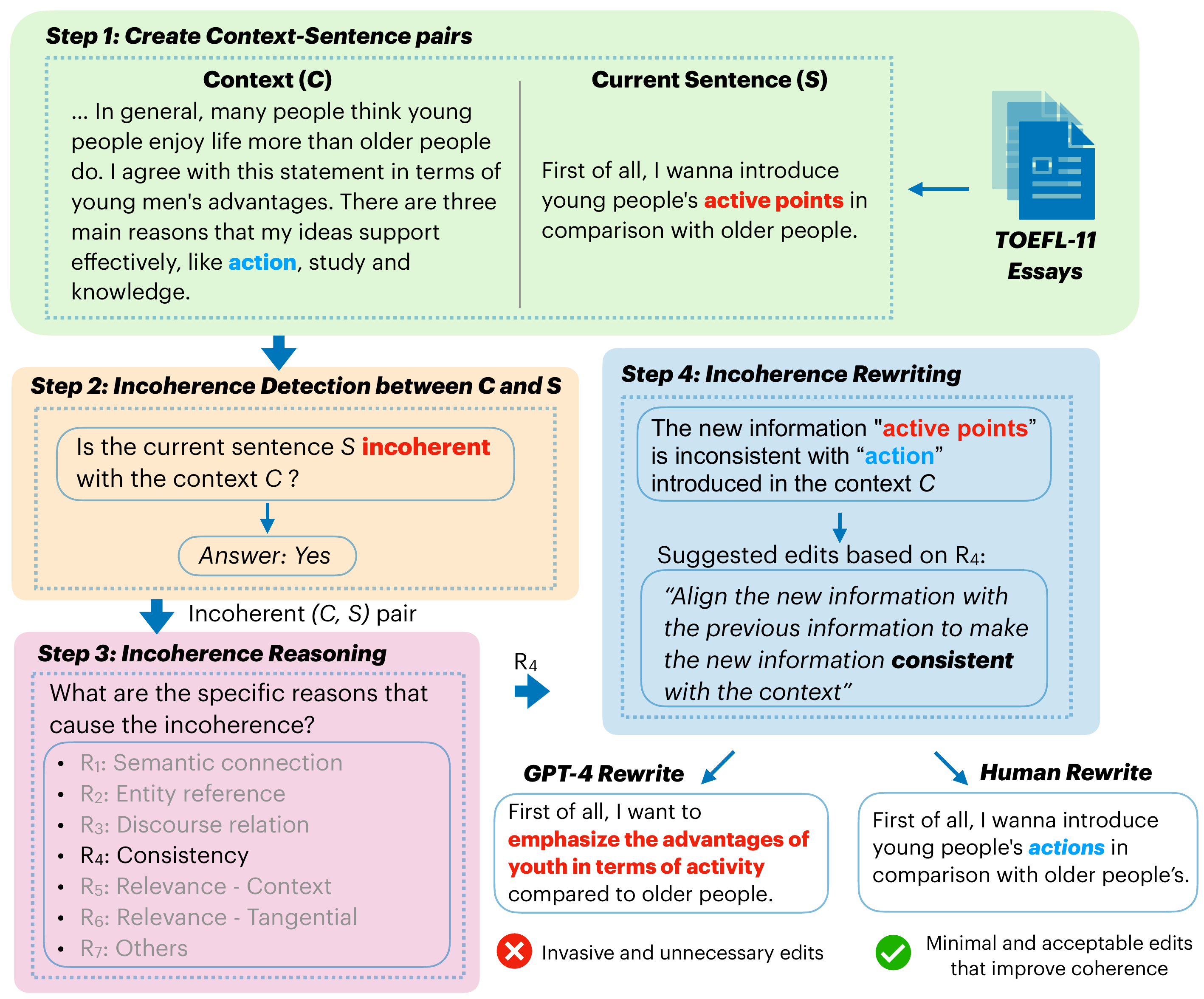}
  \caption{The overview of \textit{DECOR}, containing three tasks: incoherence detection, reasoning, and rewriting. An example human rewrite is generated for the given context-sentence pair. GPT-4 rewrite is unacceptable since it generates more invasive and unnecessary changes.}
  \label{fig:data_creation_process}
\end{figure}

Hence, we introduce \textit{DECOR}, a novel benchmark dataset that can be used to improve coherence in L2 English writing. To construct \textit{DECOR}, we start by creating context-sentence pairs from the TOEFL-11 corpus \cite{blanchard2013toefl11}, following the incremental annotation protocol suggested by \citet{maimon2023cohesentia}. We then design a language-learning-oriented annotation scheme that guides expert annotators to detect incoherence in these pairs, identify specific reasons for incoherence, and rewrite the incoherent sentences. Figure \ref{fig:data_creation_process} demonstrates the overview of \textit{DECOR} and the three tasks. To our knowledge, \textit{DECOR} is the first benchmark to feature expert annotations for incoherence detection, reasoning, and rewriting, specifically tailored for L2 English writing. The resulting parallel corpus with pairs of original incoherent sentences and their expert-revised versions, provides a valuable resource for evaluating coherence in automated writing evaluation systems.

% In the illustrative example, we compare the rewrites by GPT-4 with those produced by human experts; the revisions from GPT-4 are typically more invasive and less essential.

Moreover, while previous research demonstrated the effectiveness of using GPT-4 to assess writing coherence \cite{naismith2023automated}, challenges persist, particularly for users in developing countries \cite{bubeck2023sparks,firdaus2023utilization}. These include limited access to GPT-4, the high cost associated with its usage, and its tendency to produce overly invasive and non-essential revisions (as shown in Figure \ref{fig:data_creation_process}). Consequently, building smaller and more accessible models tailored specifically to our dataset could bring significant benefits. Hence, we develop models to automatically perform the proposed three tasks on \textit{DECOR}. The findings from our experiments indicate that our incoherence detection models deliver performance
comparable to GPT-4 in zero-shot and few-shot scenarios, despite being significantly smaller and less costly. We also demonstrate that both automatic and human evaluations affirm that fine-tuning rewriting models with specific reasons for incoherence consistently enhances their ability to produce rewrites that match the quality of those generated by human annotators.

Overall our contributions are three-fold:
\begin{itemize}
    \item We introduce a novel benchmark \textit{DECOR}, comprising 1,352 context-sentence pairs, which can serve as a valuable resource for assessing coherence in automated writing evaluation systems. 
    \item We produce the first parallel corpus that includes $213$ pairs of original incoherent sentences as well as their expert-rewritten counterparts.
    \item We fine-tuned models using task-specific synthetic data and evaluated them on \textit{DECOR}. These models achieve results comparable to GPT-4 in detecting incoherence and producing rewrites that match the quality of those generated by human experts.
    % \item We fine-tuned models with task-synthetic data and assessed them on \textit{DECOR}. The models obtain results that are comparable to GPT-4 for the incoherence detection task, and generate rewrites that match the quality of expert-rewritten sentences.
\end{itemize}

% 

% coherence detecting benchmark that evaluates learner produced texts at a sentential level on the basis of how each sentence (S) aligns semantically with the context that the writer has established up until S -1.  

%%% related work with discourse markers
% -> \cite{ru2023distributed}, \cite{nie2019dissent}, \cite{pan2019discourse}

% TODO: mention GRUEN and DiscoScore as well

% , as defined in \citet{miltsakaki-etal-2004-penn}
% \paragraph{Existing benchmark and dataset}
% -> COHESENTIA Dataset: annotation for coherence detection and reasoning (no rewrite + not focused on language learning)
% other existing dataset related to discourse markers and coherence -> \cite{sileo2019mining}, \cite{prasad-etal-2008-penn}
% \paragraph{Current Evaluation Metrics}
% -> GRUEN: metrics to evaluate linguistic quality of text generation (coherence only considers sentence order: no other semantic/syntactic rules)
% -> DiscoScore

\section{Related Work}
% We discuss related work on coherence assessment. We start by referring to what the Second Language Acquisition (SLA) literature has to say about the assessment of coherence in writing created by L2 English learners. Then we refer to the recent work for assessing coherence for machine-generated texts.

% Then, we refer to how coherence is typically approached pedagogically, and finally we introduce studies that have focused on the use of discourse markers and their role in contributing to a text’s coherence.

% \subsection{Assessment of coherence in L2 English writing}
\subsection{Definitions of coherence in English writing}

Earlier efforts at defining coherence in English, such as \citet{halliday2014cohesion}, focus on explicit cohesive ties (e.g. semantic relations between elements).  In particular, \citet{halliday2014cohesion} define cohesion as a combination of lexical and grammatical items that facilitate sentences to be understood as connected discourse rather than individual sentences. Moreover, \citet{lautamatti1978observations} defined Topical Structure Analysis (TSA) that focuses on different types of progression that are used to create coherence in a text to advance the discourse topic \cite{knoch2007little}. Additionally, \citet{reinhart1980conditions} introduced three conditions for a text to be coherent: cohesion, consistency, and relevance, capturing various aspects of the text. In developing our annotation scheme, we referred to these previous efforts and established a useful guideline that is beneficial for annotating incoherence in L2 English writing.

% previous research on coherence assessment 
% We attempted to build on top of these efforts

% TSA has become a popular method of analyzing the coherence of L2 English writing and its categories for defining different types of textual progression have been extended further through work that focused on its application to the evaluation of learner-produced texts \cite{schneider1990analyzing,knoch2007little}.

% In contrast, topic-based analysis identifies key concepts through their frequency of use in a text and line diagrams that represent the schemata of the discourse are drawn. While this method of measuring coherence has been shown to be consistent and robust to differences between inter-rater judgments, due to the complexity of this approach, it has been noted to be too complicated and time consuming and not suited to the application of statistical analysis (Knoch, 2007; Knoch, et al., 2014). In a paper by Knoch et al. (2014) that investigated the effect of task type on the discourse produced by students at various score levels of the TOEFL iBT writing test, Knoch et al. opted to use a simplified version of Lautamattti’s TSA as a measure of textual coherence in their study citing that TSA seemed to be a suitable measure of coherence for a large number of writing samples while also allowing them to use one measure for coherence in their study.

\subsection{Assessing coherence in machine-generated texts and human-written texts}
\paragraph{Machine-generated texts}
Following the linguistic definition of coherence established in \citet{reinhart1980conditions}, a more recent work by \citet{maimon2023cohesentia} incorporated these conditions into a novel benchmark, namely CoheSentia, and proposed a new coherence-annotation protocol that aligns better with human judgments. Unlike previous work that assigns a single holistic coherence score to each target text \cite{lai2018discourse}, CoheSentia provides incremental coherence labeling on a sentence-by-sentence basis, enabling humans to identify the specific reasons for incoherence. In our human annotation process, we follow the CoheSentia protocol to create the context-sentence pairs incrementally. We expand the linguistic fundamentals applied in CoheSentia and devise an annotation scheme that is tailored to incoherence detection and rewriting in L2 English writings.
\paragraph{Human-written texts} NLP techniques of
Coherence detection for human-written texts primarily identified simple surface feature proxies. \citet{mcnamara2010coh} developed Coh-Metrix that measures cohesion from a wide range of linguistic indexes. Similarly, \citet{crossley2016tool} proposed a toolkit for automatic analysis of text cohesion. Recent work by \citet{naismith2023automated} investigated the ability of GPT-4 to produce ratings for discourse coherence assessment.

\section{\textit{DECOR} Benchmark and Annotation Scheme}
\label{sec:benchmark}

In this section, we detail the data creation process for \textit{DECOR} (Section \ref{sec:data_creation}). We also outline the specific annotation schemes for each proposed task: Incoherence Detection (Section \ref{sec:ann_inc_detection}), Incoherence Reasoning (Section \ref{sec:ann_inc_reasoning}), and Incoherent Sentence Rewriting (Section \ref{sec:ann_rewriting}).

% Figure \ref{fig:data_creation_process} demonstrates the overall process of constructing \textit{DECOR}.

\subsection{Data Creation}
\label{sec:data_creation}
We propose \textit{DECOR}, a benchmark for assessing the writing coherence in L2 English essays. To construct the dataset, we first sampled $100$ medium-level essays from the TOEFL-11 dataset \cite{blanchard2013toefl11}. Note that sentences from the TOEFL-11 dataset often have basic grammar mistakes like spelling errors or missing be-verbs, which can make the intended meaning unclear. Therefore, we used a grammar error correction model from \citet{zhang2024prolex} to fix these mistakes without changing the overall meaning of the sentence. 
% and corrected their basic grammar errors using the grammar models from \citet{zhang2024prolex}. 
Then, we incrementally constructed context-sentence pairs $(C,S)$ for each essay, following the protocol suggested by \citet{maimon2023cohesentia}. In these pairs, sentence $S$ is the current sentence to be assessed, and context $C$ includes all preceding sentences in the essay up to and including the sentence immediately before $S$. Overall, we constructed $1,352$ $(C,S)$ pairs from the $100$ essays. The general statistics of \textit{DECOR} are shown in Table \ref{tab:stats_general}. 
\begin{table}[ht]
\centering
\begin{tabular}{lr}
\toprule
Items                       & Count \\ \midrule
\# of essays               & $100$   \\
\# of words                & $26,376$ \\
\# of context-sentence pairs            & $1,352$  \\
\# of coherent sentences   & $906$   \\
\# of incoherent sentences & $446$   \\
\# of human rewrites              & $213$   \\ \bottomrule
\end{tabular}
\caption{Overall statistics of \textit{DECOR}.}
\label{tab:stats_general}
\end{table}
More detailed statistics, such as the number of sentences and words per essay, are shown in Figure \ref{fig:other_stats} in the Appendix. Next, for each context-sentence pair $(C, S)$, we ask our human annotators to complete three tasks according to our annotation schemes: incoherence detection, reasoning, and rewriting. These three tasks are the main features of \textit{DECOR}. We discuss these features and their specific annotation schemes below.

% \begin{figure}[t]
%   \includegraphics[width=\columnwidth]{example-image-golden}
%   \caption{The number of sentences and words per essay in DECOR.}
%   % \label{tab:sent_word_essay}
% \end{figure}

% \caption{The number of words per rewrite generated by annotators.}

% coherence assessment in English essay writing via detection, reasoning, and sentence rewrite

\begin{table*}[t]
\centering
\begin{adjustbox}{max width=.94\linewidth}
\begin{tabular}{lll}
\toprule
\multicolumn{1}{c}{Label Codes} & \multicolumn{1}{c}{Descriptions}                                                                                                                                                              & \multicolumn{1}{c}{Examples}                                                                                                                                                              \\ \midrule
\large R1: Semantic connection              & \begin{tabular}[c]{@{}l@{}}The sentence $S$ does not connect semantically \\ with the context $C$.\end{tabular}                                                                                   & \begin{tabular}[c]{@{}l@{}}$C$: If students study ideas and concepts, they can explore new areas of research.\\ $S$: \textcolor{red}{We} need to make effort to apply \textcolor{red}{our} knowledge \\ $S'$: \textit{\textbf{They}} need to make effort to apply \textit{\textbf{their}} knowledge.

\end{tabular}               \\ \rowcolor{verylightgray}
\large R2: Entity reference                 & \begin{tabular}[c]{@{}l@{}}The current sentence $S$ discusses an entity that \\ has not been introduced in $C$ yet, or \\ sentence $S$ discusses an entity that is ambiguous in $C$.\end{tabular} & \begin{tabular}[c]{@{}l@{}}$C$: Some people enjoy tours. \\ $S$: \textcolor{red}{Guides} provide a lot of value for \textcolor{red}{tourists}. \\ $S'$:\textbf{\textit{Traveling in tour groups}} provides a lot of value for \textbf{\textit{them}}. 
\end{tabular}                                                                     \\
\large R3: Discourse relation               & \begin{tabular}[c]{@{}l@{}}The relation between sentence $S$ and previous \\ ones in $C$ doesn’t make sense due to a missing \\ discourse marker.\end{tabular}                                    & \begin{tabular}[c]{@{}l@{}}$C$ Advertisements are not good for consumers. \\ $S$: They only show the good features of a product. \\ $S'$: \textbf{\textit{For example}}, they only show the good features of a product.
\end{tabular}                                                \\ \rowcolor{verylightgray}
\large R4: Consistency                       & \begin{tabular}[c]{@{}l@{}}The current sentence $S$ contradicts or is inconsistent \\ with previously presented  information.\end{tabular}                                                          & \begin{tabular}[c]{@{}l@{}}$C$: Because gas is getting more expensive, less people will drive in the future.\\ $S$: Scientists are finding ways to make gas \textcolor{red}{cheaper} for drivers. \\ $S'$: Scientists are \textbf{\textit{researching alternative sources of energy}}. 
\end{tabular} \\
\large R5: Contextual relevance              & \begin{tabular}[c]{@{}l@{}}The current sentence $S$ introduces information that \\ is completely irrelevant to the context.\end{tabular}                                                            & \begin{tabular}[c]{@{}l@{}}$C$: To become successful, people need to take risks.\\ $S$: I think fear controls our decision making process. \\ $S'$: \textbf{\textit{Risks are important for people to learn what works and what doesn't work}}. 

\end{tabular}                                                  \\ \rowcolor{verylightgray}
\large R6: Tangential relevance            & \begin{tabular}[c]{@{}l@{}}The current sentence $S$ introduces information that \\ is tangential or unnecessary for the development \\ of the context.\end{tabular}                                 & \begin{tabular}[c]{@{}l@{}}$C$: Young people tend to not help the people of their community.\\ $S$: When I was younger I used to volunteer at a retirement home. \\ $S'$:\textbf{\textit{As a result, there may be a lack of volunteers an places like retirement homes}}. 

\end{tabular}                 \\
\large R7: Others                           & \begin{tabular}[c]{@{}l@{}} Other reasons that are not listed above. For example, \\ the comment (rheme/focus) of the sentence does \\ not agree with the  topic of the sentence.\end{tabular}                                                          & \begin{tabular}[c]{@{}l@{}}$S$: My pet fish is \textcolor{red}{flying} in the \textcolor{red}{sky}. \\ $S'$: My pet fish is \textit{\textbf{swimming}} in \textit{\textbf{its tank}}.
\end{tabular} 

\\ \bottomrule
\end{tabular}
\end{adjustbox}
\caption{Label codes for the specific reasons for incoherence during annotation. The rewrites $S'$ are provided for each incoherent $(C,S)$ pair. The erroneous parts in $S$
are marked in red, and the corrections are marked bold in $S'$.}
\label{tab:task2_desc}
\end{table*}

\subsection{Incoherence Detection Annotation Scheme}
\label{sec:ann_inc_detection}
% \textit{DECOR} features the ability to detect the incoherence of a given context-sentence pair. 
Inspired by the linguistic fundamentals of coherence (i.e. cohesion, consistency, and relevance) defined in \citet{reinhart1980conditions}, we expanded these fundamentals with reference to previous work in order to apply the task of incoherence detection to L2 English writing. We describe five specific criteria for detecting incoherence in each context-sentence pair below.

% the features of DECOR's annotation scheme for incoherence detection below. 

\textbf{Semantic connection} serves as the criterion that is based on the expanded categories of discourse progression for TSA proposed in \citet{lautamatti1978observations}, where a sentence’s semantic connection with the context of discourse is defined by its appropriate use of the sequential progression of topics from sentence to sentence that contributes to local coherence (\citealt{reinhart1980conditions,knoch2007little}). \textbf{Entity reference} refers to the requirement for writers to establish a link between the topics of the current sentence and the context of the discourse and is related to cohesion. Accurate anaphoric pronominal use is a key component of this criterion \cite{knoch2007little}. For instance, in the passage \textit{Learning about \textit{ideas and concepts} is essential for all students. For example, \textit{they} help students to apply their knowledge in new ways.}, the pronoun \textit{they} in the second sentence agrees in person and number with the referent \textit{ideas and concepts} in the first sentence. \textbf{Discourse relation} is concerned with how the sentence is related to the overall context through the use of explicit cohesive ties that refer to the semantic relations between an element in a text and some other element that is crucial to the interpretation of it \cite{halliday2014cohesion}. 
\textbf{Consistency} is associated with the logical requirements for a sentence to align with the preceding sentences in the context \cite{maimon2023cohesentia}.
\textbf{Relevance} dictates a sentence must be related to previous sentences in the discourse and the underlying discourse topic of the global context \cite{maimon2023cohesentia}.

% Knoch’s (2007) expanded categories of discourse progression for TSA,

If the given context-sentence pair violates any of the aforementioned criteria, it is considered incoherent, necessitating the subsequent step (described in Section \ref{sec:ann_inc_reasoning}) to identify the specific reasons causing sentence $S$ to be incoherent to context $C$; otherwise, the sentence is labeled as coherent. The detailed annotation guidelines for this task are demonstrated in Appendix \ref{app:annotation_scheme}. Note that annotators are instructed to evaluate the entire context when determining if a sentence is incoherent. If the context pertains to a fictional setting, such as a dream about fiction, these instances will not be considered incoherent.

% Note that annotators are asked to consider the whole context to judge if the sentence is incoherent. If the context relates to a fictional setting (e.g. dreaming about fiction), then we will not count it as incoherent instances. 

% The coherent $(S, C)$ pairs are marked $1$, whereas incoherent ones are marked with $-1$. For other cases that have nothing to do with writing coherence (e.g. sentence parsing errors), $0$ is marked.

\subsection{Incoherence Reasoning Annotation Scheme}
\label{sec:ann_inc_reasoning}
In addition to detecting incoherence, annotators are tasked with identifying the specific reasons for incoherence in the context-sentence pairs that are labeled as such. Drawing on the linguistic principles of coherence outlined in \citet{reinhart1980conditions}, three primary factors contribute to incoherence: \textit{Cohesion}, \textit{Consistency}, and \textit{Relevance}. Given that \textit{Cohesion} pertains to the linear sequencing and connections of sentences, we specifically designated three label codes for annotations within this category: semantic connection, entity reference, and discourse relation. For \textit{Consistency}, we use a single code: consistency. Regarding \textit{Relevance}, we have devised two codes: contextual relevance and tangential relevance. Other possible reasons that are not listed above are referred to as others. Detailed descriptions and examples of each label code are illustrated in Table \ref{tab:task2_desc}.

\subsection{Incoherent Sentence Rewriting Annotation Scheme}
\label{sec:ann_rewriting}
After selecting all applicable reasons, sentence $S$ is rewritten by the annotators to convert it to be coherent with context $C$. Concretely, annotators are asked to make the least invasive changes necessary to improve the coherence based on the identified reasons. For example, if \emph{Discourse relation} is selected as the reason, annotators are instructed to \emph{add or change a discourse marker that ties sentence S with context C}. The complete list of suggested edits is described in Appendix~\ref{app:edits}. Considering the challenges of providing all possible edits to sentence S during the annotation process, we instructed our annotators to provide only one possible edit that addresses at least one selected reason from the previous step. We leave the exploration of multiple edits for future work.

% \subsection{Data Creation}
% We sampled $100$ essays from medium-level essays in TOEFL-11 \cite{blanchard2013toefl11}.

% \subsection{DECOR Annotation Scheme}

\section{Data annotation process and statistics}
\label{sec:ann_process_statistics}

Considering the need for substantial experience in English essay grading, we recruited two annotators with extensive teaching experience in English and advanced degrees in Applied Linguistics, specializing in English language education. Before annotating \textit{DECOR}, we conducted a tutorial session to train the two annotators and familiarize them with our annotation scheme. Subsequently, in accordance with our specified scheme, we tasked them with annotating five sample essays, which comprised $72$ sentence-context pairs. 

We calculated the inter-annotator agreement for these pairs using Cohen's Kappa \cite{cohen1960coefficient}. The two annotators achieved a $\kappa$ value of $0.83$ for Incoherence Detection, indicating an almost perfect agreement. For Incoherence Reasoning, they reached an average $\kappa = 0.90$ across all reason types, also reflecting almost perfect agreement. The specific agreement scores for each reason type and more justifications for the annotation process are presented in Appendix~\ref{app:iaa_reason_type}. As for Incoherent Sentence Rewriting, the leading authors validated whether the new sentences are acceptable. In particular, a new sentence $S'$ is acceptable if it preserves the semantic meaning of the original sentence $S$ and is coherent with the given context $C$. Overall, the rewrites by the two annotators were deemed acceptable at rates of 88\% and 89\%, respectively.

% Overall, $88\%$ and $89\%$ of the rewriting from the two annotators were deemed acceptable, respectively. 

% Subsequently, the two annotators work on the complete test set separately, with each annotating $50$ essays that consist of $720$ and $734$ $(S,C)$ pairs, respectively.

% \begin{figure}[t]
%   \includegraphics[width=\columnwidth]{example-image-golden}
%   \caption{Human as a judge evaluation results with GP4 as the baseline.}
%   \label{fig:dist_incoh_reasons}
% \end{figure}

\begin{figure}[ht]
    \centering
    \begin{subfigure}[b]{0.4\textwidth}
        \centering
        \includegraphics[width=\textwidth]{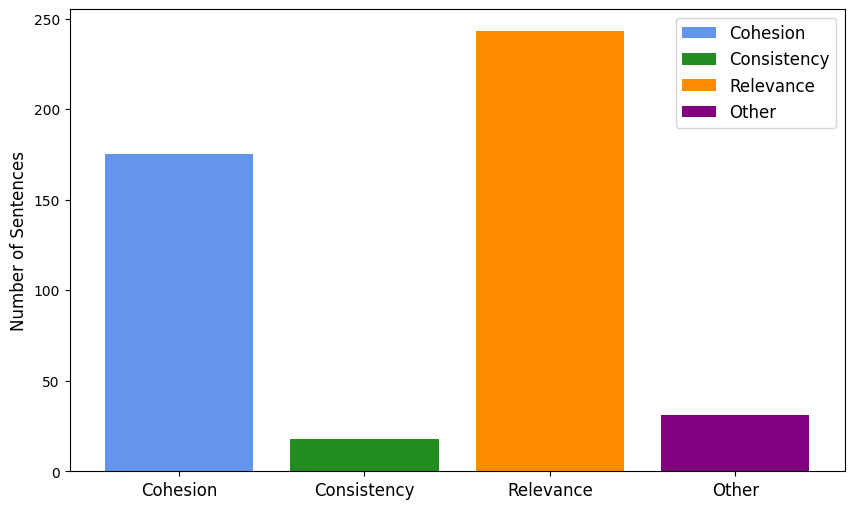}
        \caption{Distribution of reasons for incoherence clustered into groups.}
        \label{fig:dist_inc_reasons_high}
    \end{subfigure}

    % Add some vertical space between the subfigures
    \vspace{0.2cm}

    \begin{subfigure}[b]{0.4\textwidth}
        \centering
        \includegraphics[width=\textwidth]{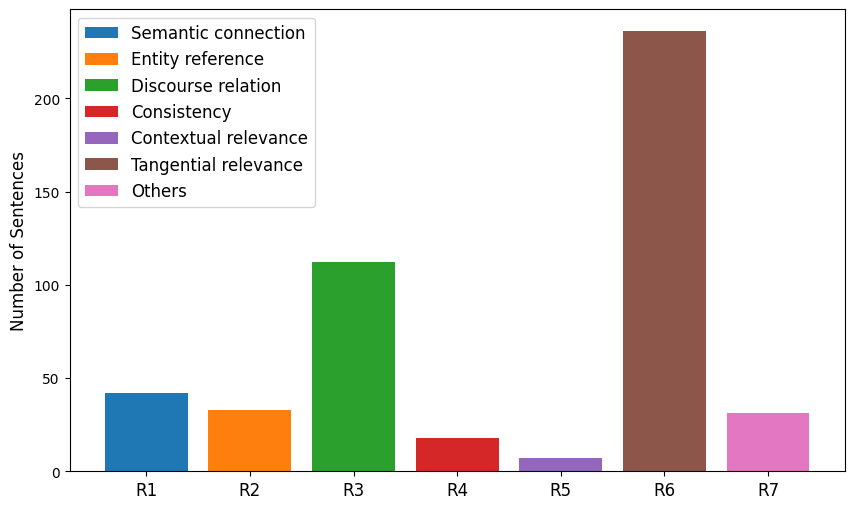}
        \caption{Distribution of specific reasons for incoherence.}
        \label{fig:dist_inc_reasons}
    \end{subfigure}

    \caption{Distribution of specific reasons for incoherence, and those clustered into groups.}
    \label{fig:dist_reason_types}
\end{figure}

Subsequently, the two annotators worked independently on the test set, with each annotating around $700$ $(C,S)$ pairs that are constructed from Section \ref{sec:benchmark}. Overall, among all $1,352$ $(C,S)$ pairs, $906$ sentences are coherent with their corresponding contexts, whereas $446$ sentences are labeled as incoherent. We present the number of words per rewrite in Figure \ref{fig:words_rewrite}. Note that we do not consider rewrites marked as \textit{DELETE},  resulting in $213$ rewrites that contain more than one word. In addition, we presented the distribution of the annotated reasons for incoherence in Figure \ref{fig:dist_reason_types}. Our analysis shows that the medium-level essays, randomly sampled from the TOEFL-11 corpus, generally maintain consistency and rarely contradict the context. Moreover, we also find that the primary sources of incoherence in these essays are related to \textit{Relevance} and \textit{Cohesion}, with issues of tangential relevance and weak discourse relations being the most prevalent.

\begin{table*}[t]
\centering
\begin{adjustbox}{max width=.7\textwidth}
\begin{tabular}{rl|ccccc}
\toprule
\multicolumn{2}{c|}{\multirow{2}{*}{\textbf{Models}}} & \multicolumn{1}{c|}{\multirow{2}{*}{\textbf{\begin{tabular}[c]{@{}c@{}}Incoherence \\ Detection (\%)\end{tabular}}}} & \multicolumn{4}{c}{\textbf{Incoherence Reasoning (\%)}} \\ \cmidrule{4-7} 
\multicolumn{2}{c|}{} & \multicolumn{1}{c|}{} & \textit{Cohesion} & \textit{Consistency} & \textit{Relevance} & \textit{Others} \\ \midrule

\multirow{2}{*}{BERT-base} & $D_C$ & 63.04 & 48.17 & 93.76 & 28.47 & - \\
                           & $D_T$ & 66.43 & 44.38 & 75.41 & 46.37 & {80.36} \\ \cmidrule{2-7}
\multirow{2}{*}{DeBERTa-base} & $D_C$ & 62.21 & 47.93 & \textbf{93.88} & 29.45 & - \\
                             & $D_T$ & {68.54} & 48.36 & 77.17 & 45.14 & 74.20 \\ \midrule
\multirow{2}{*}{Llama2-7B} & $D_C$       & 59.52 & 43.93 & 93.65 & 28.87 & - \\
                           & $D_T$       & 66.08 & 46.63 & 83.55 & 47.20 & 87.78 \\ \cmidrule{2-7}
\multirow{2}{*}{GPT-4}     & $zero$ & 66.56 & \textbf{51.03} & {93.02} & 56.60 & \textbf{87.93} \\
                           & $16$ & \textbf{69.33} & {48.71} & 90.10 & \textbf{65.54} & 85.64 \\ \bottomrule
\end{tabular}
\end{adjustbox}
\caption{Evaluation of models on \textit{DECOR} using weighted F1 scores in percentages (\%) for Incoherence Detection and Incoherence Reasoning tasks. For each task, the task-specific synthetic training data is denoted as $D_T$, whereas the out-of-domain training data is denoted as $D_C$. We also conducted zero-shot ($zero$) and in-context learning (16-shot) with GPT-4. Since \textit{Others} is not specified in $D_C$, we exclude it for evaluation.}
\label{tab:eval_task12}
\end{table*}

\section{Incoherence Detection, Reasoning and Rewriting}
% Ablation: Evaluate sentence rewrite w VS w/o reasoning selection
% -> Compare models' ability to rewrite incoherent sentences with reasons VS w/o reasons (in training/inference) -> human eval

We propose \textit{DECOR} to benchmark the model's ability in incoherence detection, reasoning, and rewriting for English essays written by L2 language learners. In this section, we will outline each of the three tasks and describe their specific task formulations, evaluation metrics, data, baselines, and results and analysis.

\subsection{Incoherence Detection}
\subsubsection{Task formulation}

In this task, the model will assess the given context-sentence pairs that are extracted from essays written by L2 learners, determining whether the sentence $S$ maintains coherence with the context $C$. This task is specifically designed to evaluate the effectiveness of systems in capturing coherence within learner-written texts.

% whats the task?
% how many data points does the task have?
% what are the evaluation metric
% what are the baselines
% what are the results

\subsubsection{Evaluation metrics}

Given the class imbalance in our test set, where $906$ instances are labeled as coherent and $446$ as incoherent, we opt to use the weighted F1 score as a metric to assess the performance of different models. This approach ensures a fair evaluation by accounting for the disproportionate distribution of classes.

% Considering the class imbalanced issue in our test set (i.e. 906 labeled as coherent while 446 labeled as incoherent), we adopted the weighted F1 score to evaluate the results from the models.

% We evaluate the results from the models with 

\subsubsection{Data}
\label{sec:syn_data_task1}

% synthetic training data
Given the absence of a dedicated incoherence detection corpus for language learners suitable for model training purposes, we followed the approach recommended by \citet{zhang2024prolex} and synthesized task-specific incoherence detection data using GPT-4 \cite{openai_gpt4}.\footnote{Throughout this paper, we employ \texttt{GPT-4o} as the default model unless otherwise specified.} The prompt we used for GPT-4 is shown in Appendix~\ref{app:syn_prompt_gpt}. To start with, we randomly sampled $800$ medium-level essays from the TOEFL-11 dataset and generated $11,267$ context-sentence pairs. We then used GPT-4 to analyze these pairs for incoherence, producing a label for each. In this process, $6,422$ sentences were identified as coherent, while $4,845$ were labeled as incoherent. For the training process, we allocated $90\%$ of this synthetic data for training purposes, denoted as $D_T$, and reserved the remaining $10\%$ for validation. Moreover, we also utilized out-of-distribution (OOD) training data proposed in \citet{maimon2023cohesentia}, denoted as $D_C$. To our knowledge, $D_C$ is the only dataset featuring human annotations for coherence detection and incoherence reasoning in machine-generated texts. We incorporated $D_C$ as OOD training data to assess whether models trained exclusively on it could achieve strong performance in \textit{DECOR}. This experiment aims to determine whether using OOD data can improve the detection of incoherence in texts authored by L2 speakers.

% \footnote{Note that each task has its own $D_T$ and $D_C$. We use them to represent the concept of the task-specific synthetic training data and the out-of-distribution data, respectively.}

% With $D_T$ and $D_C$, we trained and evaluated two classes of models: classification-based and generation-based.

\subsubsection{Baselines}

We conducted experiments with classification-based models that consist of encoder-only architectures equipped with a classification head. Specifically, we tested models such as BERT \cite{devlin2018bert} and DeBERTa \cite{he2021debertav3} with their base and large variants. Each model generates predictions with two labels—\textit{yes} or \textit{no}—to determine if the sentence $S$ is coherent with the context $C$. The input to the model's encoder is structured in the format "$C$ <SEP> $S$," facilitating the assessment of coherence between the given context and sentence.

In light of the burgeoning field of powerful instruction-following models \cite{ouyang2022training,bai2022training,openai_gpt4}, we also explored two generation-based large language models: Llama 2 \cite{touvron2023llama} and GPT-4. For Llama 2, we fine-tuned its 7B variant using our synthetic dataset $D_T$ for this task. With GPT-4, we tested in both zero-shot and 16-shot settings. Details of the prompts used in the GPT-4 experiments are provided in Appendix~\ref{app:gpt4_all_prompts}. 

% We experimented the classification-based models, which are encoder-only models with a classification head. Specifically, we tested with BERT \cite{devlin2018bert} and DeBERTa \cite{he2021debertav3}. The output of each prediction is 2 labels (i.e. yes or no) to predict if the sentence $S$ is coherent with context $C$. The input to the encoder of the models is formulated as "$C_i$ <SEP> $S_i$".

\subsubsection{Results and analysis}
\label{sec:task1_results}
% training with D_T is better than training with D_C for -> bert close to zero-shot gpt-4 result, deberta surpass gpt-4 zero shot and almost the same as 16-shot
% 

The results for the task of incoherence detection are demonstrated in Table \ref{tab:eval_task12}. As observed, training with our task-specific synthetic dataset $D_T$ yielded superior results compared to using the OOD dataset $D_C$. This improvement is attributed to the fact that $D_C$ consists solely of machine-generated texts, which introduces a significant distribution shift. Additionally, while GPT-4 with 16-shot examples surpassed all other models, smaller models trained on our synthetic data $D_T$, such as BERT-base and Llama-2-7B, achieved performance comparable to GPT-4 in a zero-shot setting. Moreover, DeBERTa-base matched GPT-4's performance in the 16-shot setting and even exceeded it in the zero-shot scenario. We also experimented with combining both $D_C$ and $D_T$ during training; however, this did not lead to improved results. Details of the experiment are provided in Appendix~\ref{app:details_baseline}.

% outcomes of these experiments are provided in Table \ref{tab:baseline-app}. 

\subsection{Incoherence Reasoning}
\subsubsection{Task formulation}
\label{sec:task_form_inc_reason}
The incoherence reasoning task aims to develop models capable of identifying the specific causes of incoherence in context-sentence pairs labeled as such. Due to the sparse distribution of incoherence reason types depicted in Figure \ref{fig:dist_inc_reasons}, we focus on the four high-level causes previously introduced: \textit{Cohesion}, \textit{Consistency}, \textit{Relevance}, and \textit{Others}. For each of these four causes of incoherence, we developed specialized models capable of determining whether the incoherence stems from a specific cause. This approach divides the overall incoherence reasoning task into four distinct sub-tasks, each targeting a different cause.

% Specifically, drawing on the insights from the previous research \cite{maimon2023cohesentia}, we categorized the seven reasons for incoherence, as annotated in Section \ref{sec:inc_reason}, into four high-level causes: \textit{Cohesion} (including semantics connection, entity reference, and discourse relation), \textit{Consistency}, \textit{Relevance} (i.e. context and tangential), and \textit{Others} (i.e. topic-comment-disagreement). 

% For example, in detecting \textit{Cohesion} as the cause, an instance label is predicted as "Yes" if either 

% in creating the training set for detecting \textit{Cohesion} as the cause, an instance is labeled "Yes" if GPT-4 identifies R1, R2, or R3 as the cause of incoherence for that instance; otherwise, the label is "No", indicating that the incoherence is caused by other factors.

% , the goal of this task is to accurately identify the factors contributing to incoherence in any sentence $S$ that lacks coherence with its context $C$. 

\subsubsection{Evaluation metrics}

In Figure \ref{fig:dist_inc_reasons_high}, \textit{DECOR} exhibits class imbalance across the four reason types of incoherence. Hence, we report weighted F1 scores for each of the four sub-tasks to account for this imbalance.

% As depicted in Figure \ref{fig:dist_inc_reasons_high}, \textit{DECOR} features unbalanced annotations for the four reason types of incoherence. Hence, we report the weighted F1 scores for each of the four sub-tasks to account for this imbalance.

% the test set features unbalanced annotations for the \textit{Consistency} and \textit{Other} categories. Consequently, we report the weighted F1 scores for these two sub-tasks to account for this imbalance. In contrast, for the \textit{Cohesion} and \textit{Relevance} categories, which have relatively balanced annotations in the test set, we utilize standard F1 scores for evaluation.

% the test set contains unbalanced annotations for \textit{Consistency} and \textit{Other}. Hence, we report the weighted F1 scores for the two sub-tasks. In contrast, we use the standard F1 scores to evaluate \textit{Cohesion} and \textit{Relevance}, given their relatively balanced annotations in the test set.

% Among the $446$ context-sentence pairs that are labeled as incoherent, 

\subsubsection{Data}
\label{sec:task2_data}
% synthesize data for reasoning selection for each sub-task

We adopted a similar approach as described in Section \ref{sec:syn_data_task1} to synthesize the training data for all four sub-tasks. Specifically, we prompted GPT-4 to identify all potential reasons for each instance of incoherence detected from Section~\ref{sec:syn_data_task1}, based on the seven predefined causes outlined in Table~\ref{tab:task2_desc}. The prompts we used for data synthesis are demonstrated in Appendix~\ref{app:syn_prompt_gpt}. Furthermore, we post-processed the resulting data to create four distinct datasets, each serving as the training data for detecting \textit{Cohesion}, \textit{Consistency}, \textit{Relevance}, and \textit{Others}. For instance, in creating the training set for detecting \textit{Cohesion} as the cause, an instance is labeled "Yes" if GPT-4 identifies R1, R2, or R3 as the cause of incoherence for that instance; otherwise, the label is "No", indicating that the incoherence is caused by other factors. Similar to \ref{sec:syn_data_task1}, the synthetic datasets are denoted as $D_T$. The details for the post-processing and statistics of the resulting data for each sub-task are described in Appendix~\ref{app:syn_data_postprocessing}. 

% In addition, we also experimented with the out-of-distribution training data that was used in \citet{maimon2023cohesentia} for each categorical reason type, denoted as $D_C$.

% decompose the resulting synthesized data into four sub-sets for the four sub-tasks as defined above in Section 

\subsubsection{Baselines}

We adopted the same set of baseline models that are tested in the incoherence detection task: classification-based models (i.e. BERT and DeBERTa), and generation-based models (i.e. Llama 2 and GPT-4). Similarly, for each sub-task of the incoherence reasoning, each model predicts with two labels (i.e. yes or no) to determine if the sentence S is incoherent with the context C due to a specific cause. We fine-tuned BERT, DeBERTa, and Llama2-7B models on the task-specific synthetic data $D_T$ for each sub-task as well as the out-of-distribution data $D_C$. We also prompted GPT-4 under both zero-shot and 16-shot settings. The prompts for GPT-4 experiments are shown in Appendix~\ref{app:gpt4_all_prompts}.

% both BERT and DeBERTa models with the task-specific synthetic data $D_T$ for each sub-task as well as the out-of-distribution data $D_C$. We fine-tuned Llama 2 with $D_T$ for each sub-task, and prompted GPT4 under both zero-shot and 16-shot settings. The prompts for GPT-4 experiments are shown in Appendix~\ref{app:gpt4_all_prompts}.

\subsubsection{Results and analysis}
% training with D_T helps - consistency and others sucks for D_C
% bert and deberta did not perform well in cohesion and relevance, even with D_T -> hard task (even gpt-4 does not reach to 60% for cohesion)
% GPT-4 outperforms all baselines, Llama2-7B beats zero-shot in cohesion, and is comparable in relevance and others
% suggest limitations for the imbalanced dataset -> increase the size in future work

The results for incoherence reasoning in terms of the four sub-tasks are demonstrated in Table \ref{tab:eval_task12}. It was observed that training DeBERTa-base and Llama2-7B models with $D_T$ resulted in enhanced performance for \textit{Cohesion} and \textit{Relevance} when compared to training with $D_C$. For \textit{Cohesion}, DeBERTa-base outperforms the Llama2-7B model and is close to the performance of GPT-4. In comparison, for the \textit{Consistency} task, all of our models demonstrate markedly enhanced performance when trained with $D_C$ rather than $D_T$. This improvement is likely attributed to the imbalanced training data distribution in $D_C$, which more closely mirrors the \textit{Consistency} class distribution in \textit{DECOR}. For the task of \textit{Others}, we have omitted $D_C$ from the table because the category \textit{Others} is not included in $D_C$. Our Llama2-7B model, fine-tuned with $D_T$, achieved results comparable to GPT-4 in both zero-shot and 16-shot settings. We further explored the effects of combining $D_T$ and $D_C$ as training data to fine-tune our models for tasks excluding \textit{Others}. The results varied across different tasks and are presented in Table \ref{tab:baseline-app} in Appendix \ref{app:details_baseline}.

% Additionally, our Llama2-7B model, fine-tuned with $D_T$, achieved results comparable to GPT-4 in both zero-shot and 16-shot settings across all sub-tasks, except for the \textit{Consistency} task. Given the limited size of the test data for the \textit{Consistency} and \textit{Others} tasks, we plan to expand the test set in future work to facilitate a more balanced comparison among different models.

% training with task-specific synthetic data $D_T$ yielded better results across all four sub-tasks compared to training with out-of-distribution data $D_C$, with particularly significant improvements in the \textit{Consistency} and \textit{Others} tasks. 

% In addition, our Llama2-7B model fine-tuned with $D_T$ achieved comparable results with GPT-4 under both zero-shot and 16-shot settings for all sub-tasks except for the \textit{Consistency} task. Considering the small set of test data for \textit{Consistency} and \textit{Others}, we will work on expanding the test set in future work to ensure a more balanced comparison among different models.

\subsection{Incoherence Rewriting}
\label{sec:incoherence_rewriting}
\subsubsection{Task formulation}
% \paragraph{The incoherent sentence rewriting task} 

The incoherence rewriting task is designed to assess the model's capability to edit a given incoherent sentence $S$ to a revised sentence $S'$ that restores the coherence with context C, based on the identified reasons $R$ for incoherence. Specifically, we prefer edits that not only enhance the coherence of the original sentence but also minimize alterations, ensuring the changes are as unobtrusive as possible.

\begin{table}[t]
\centering
\begin{adjustbox}{max width=\columnwidth}
\begin{tabular}{lrrr}
\toprule
\multirow{2}{*}{Model} & \multirow{2}{*}{\begin{tabular}[c]{@{}r@{}}Training\\Condition\end{tabular}} & \multirow{2}{*}{\begin{tabular}[c]{@{}r@{}}Acceptance\\Rate (\%)\end{tabular}} & \multirow{2}{*}{\begin{tabular}[c]{@{}r@{}}Win\\Rate (\%)\end{tabular}} \\
&&&\\\midrule 
\multirow{2}{*}{Llama2-7B} & w/ reason           & 75.59                               & 69.16                        \\
& w/o reason  & 74.65                               & 69.01                        \\ \midrule 
\multirow{2}{*}{Llama3-8B-Instruct} & w/ reason & \textbf{77.46} & \textbf{72.30} \\
& w/o reason &  75.12                               & 71.83                        \\   \bottomrule
\end{tabular}
\end{adjustbox}
\caption{Automatic evaluation of models for the incoherence rewriting task. The win rate is calculated by adopting GPT-4 as a judge to compare the system-generated rewrites against human-written references.}
\label{tab:task3_auto_eval}
\end{table}

\subsubsection{Evaluation metrics}

We measured the systems' performance on the task of incoherence rewriting with the acceptance rate. This metric was determined by calculating the proportion of revised sentences $S'$ that both achieve coherence with context $C$ and maintain minimally invasive edits, out of all evaluated incoherent context-sentence pairs. We specifically employed GPT-4 with 16-shot examples (with the best performance in the incoherence detection task) to determine if the rewrites $S'$ are acceptable. Additionally, in line with the recent practices of evaluating instruction-following LLMs \cite{zhou2024lima,dubois2024alpacafarm}, we asked GPT-4 to rank a pair of generated rewrites (one from the human-written reference, the other from the tested models) to decide which one is more coherent to the context $C$. For each tested model, we collect its win rate against the human reference. Note that we randomly shuffle the ordering of the pair-wise outputs to avoid position biases. The prompt we adopted for GPT-4 judging is shown in Appendix \ref{sec:gpt_judge_prompt}.

% Specifically, we present a rewrite generated by the model to be evaluated alongside the rewrite created by our human annotators. We then use GPT-4 as a judge to conduct the pair-wise comparisons between the system outputs and the human-written references, and assess 
% assess which of the two rewrites is more coherent.

\subsubsection{Data}
Given the reasons generated from the incoherence reasoning task, we prompted GPT-4 to generate the rewrites based on the identified reasons for incoherence. These rewrites are used as the training data for the incoherence rewriting task. The prompt we used for the rewrite synthesis and relevant statistics are shown in Appendix~\ref{app:syn_prompt_gpt}. For automatic evaluation, we used all $213$ rewrites generated by our annotators, and we randomly selected a sample of $100$ for human evaluation.

\begin{figure}[t]
\centering
  \includegraphics[width=.95\columnwidth]{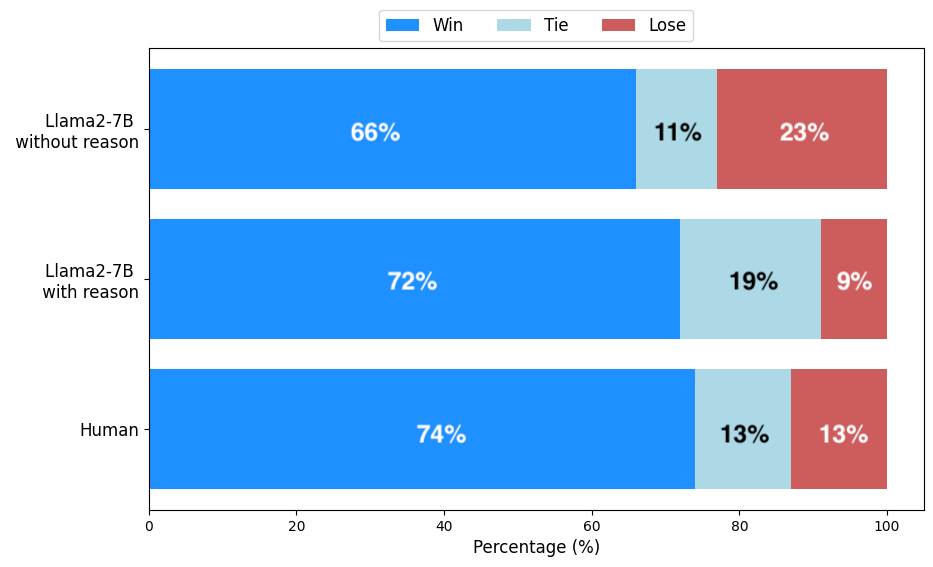}
  \caption{Human expert as a judge evaluation results with GPT-4 rewrites as the baseline. We sample 100 examples and ask our human expert for each pair of comparisons. A higher win rate and a lower loss rate indicate superior quality.}
  \label{fig:human_eval_task3}
\end{figure}

\subsubsection{Baselines}

We conducted experiments with two advanced open-sourced generative LLMs, Llama 2 \cite{touvron2023llama} and Llama3 \cite{llama3modelcard}, for the incoherence rewriting task. Specifically, we fine-tuned {Llama2-7B} and {Llama3-8B-Instruct} using our synthetic rewriting dataset under two experimental conditions: training with reasons for incoherence and without reasons.

% We experimented with two powerful open-sourced generation-based LLMs, namely Llama 2 and Llama3 \cite{llama3modelcard} for the incoherence rewriting task. We specifically fine-tuned \texttt{Llama-2-7B} and \texttt{Llama3-8B-Instruct} on our task-specific synthetic data $D_T$, under two experimental settings: with reasons and without reasons.

 % To determine whether the identified reasons for incoherence are helpful in rewriting incoherent sentences, we conducted experiments under two training settings: training with reasons and training without reasons. These settings are denoted with $wr$ and $nr$, respectively. Additionally, we conducted both zero-shot and few-shot experiments with GPT-4 on all three tasks.

\subsubsection{Results and analysis}

\paragraph{Automatic Evaluation}
The automatic evaluation results for incoherence rewriting are shown in Table \ref{tab:task3_auto_eval}. As observed, fine-tuning both the Llama2-7B and Llama3-8B-Instruct models with reasons for incoherence consistently results in better performance compared to their counterparts trained without such reasons, achieving higher scores in both acceptance rate and win rate. Table \ref{tab:qual_examples_task3} demonstrates the qualitative comparisons among example rewrites produced by our fine-tuned models.
\paragraph{Human Evaluation}
Moreover, we conducted a human evaluation where we asked our human expert to judge and compare system-generated rewrites with those produced by GPT-4.\footnote{To avoid biases, instead of the same annotators, we asked one of our leading authors to conduct the human evaluation.} The detailed information for the human evaluation process is shown in Appendix \ref{app:human_eval_d}.
% including detailed information/instructions on the evaluation criteria and specific feedback from human evaluators can make it better.
Additionally, the human evaluator was also tasked with a pairwise comparison between human-written references and the same set of GPT-4 rewrites. The results are shown in Figure \ref{fig:human_eval_task3}. As expected, our human judges predominantly preferred rewrites produced by human experts over those generated by GPT-4, with the highest win rate reaching 74\%. Consistent with the results in Table \ref{tab:task3_auto_eval}, fine-tuning Llama 2 with reasons for incoherence resulted in a higher win rate and a significantly lower loss rate compared to fine-tuning without reasons. A chi-square test indicates a significant difference between these two conditions (with p-value < 0.01). This supports our hypothesis that rewriting incoherent sentences with an understanding of their underlying causes produces higher-quality rewrites.

\section{Conclusion and Future Work}
We propose a novel benchmark \textit{DECOR} aiming to assess and improve coherence in L2 English writing. Specifically, \textit{DECOR} contains three tasks: incoherence detection, reasoning, and rewriting. Our annotation scheme allows us to produce a corpus comprising $1,352$ context-sentence pairs with coherence labels, as well as the first parallel corpus featuring $213$ pairs of original incoherent sentences and their expert-rewritten counterparts. Additionally, we fine-tuned various models with task-specific synthetic data, achieving results comparable to GPT-4 in coherence detection and generating rewrites favored by both automatic and human evaluations. In future work, we plan to enhance \textit{DECOR} by expanding its size and quality, ensuring more balanced reasoning types and multiple edits for each incoherent context-sentence pair. This enhancement will create a more comprehensive evaluation set for incoherence detection and correction, specifically tailored to L2 writing.

% In future work, we plan to expand \textit{DECOR} in both size and quality, ensuring more balanced reason type labels and multiple edits for each incoherent context-sentence pair. This will create a more comprehensive evaluation set for coherence assessment tailored to L2 writing. 

\section{Limitations}
While our benchmark may contribute to building the systems that can improve the coherence in L2 English writing, there were a number of limitations to our study.

First, the distribution of incoherence reason types is unbalanced, with the \textit{Consistency} category containing the fewest annotations among the four high-level reason types. This is due to the fact that medium-level essays from the TOEFL-11 corpus, the source of all context-sentence pairs, generally maintain consistency and seldom contradict the context. We leave our future work to diversify and balance the reason types in \textit{DECOR}, potentially by including low-level essays written by English L2 learners. 

Additionally, the texts sampled from the TOEFL-11 corpus for synthesizing our training data were limited by the specific writing prompts they addressed. This limitation may hinder the system's ability to detect coherence in learner-produced writing that responds to out-of-domain prompts not included in the TOEFL-11 corpus. Future extensions of our work include incorporating other L2 English writing corpora.

Finally, regarding the general design of our annotation scheme for coherence detection, we considered all sentences in the context up until the target sentence. However, as we found during our annotation tutorial session, sometimes issues of coherence occur due to the structuring of information that is contained in sentences that come later in the text. Future work might focus on these specific types of coherence breaks and their prevalence in L2 writing.

% Next, the edited sentences that comprise our parallel corpus tended to contain the addition of a discourse marker (e.g., for example, furthermore, therefore, etc.) in order to make the sentence coherent. However, the annotators only used a limited number of these markers, which may have had an effect on the training of the model. In future work, we plan to provide annotators with a comprehensive list of discourse markers and their functions for annotators to include in their sentence edits.

\section{Ethics Statement}

\paragraph{Reproducibility}
In this work, we utilized GPT-4 to synthesize our task-specific training data for coherence detection, reasoning, and rewriting. We also used it during the evaluation. To facilitate the reproducibility of our data synthesis process and evaluation results, we included all relevant prompts that were used in our paper. In addition, all the other models used in this research, are publicly available in peer-reviewed articles and referenced in this paper. All datasets, including our synthetic fine-tuning dataset and the annotated test set, are released.

\paragraph{Biases} 
We did not explicitly handle any bias that exists in the pre-trained language models we experimented with in this paper.

\paragraph{Human Annotators}
Both annotators were specifically recruited from the linguistics department, and they are both associate professors with extensive experience in teaching English as a foreign language and have advanced degrees in Applied Linguistics. They were paid at a rate of \$12 per hour. To protect privacy and anonymity, contributors' personal and demographic information was not collected.

\section{Acknowledgement}
We would like to thank Yanda Chen, Ryan Shea, and people from the Columbia NLP group for their valuable discussions on the paper. We also thank all reviewers for their constructive feedback and suggestions, which significantly improve our work. In addition, we extend our gratitude to our expert annotators for their time and contributions to the completeness of the benchmark.

\bibliography{custom}

\begin{thebibliography}{34}
\providecommand{\natexlab}[1]{#1}

\bibitem[{AI@Meta(2024)}]{llama3modelcard}
AI@Meta. 2024.
\newblock \href {https://github.com/meta-llama/llama3/blob/main/MODEL_CARD.md} {Llama 3 model card}.

\bibitem[{Bai et~al.(2022)Bai, Jones, Ndousse, Askell, Chen, DasSarma, Drain, Fort, Ganguli, Henighan et~al.}]{bai2022training}
Yuntao Bai, Andy Jones, Kamal Ndousse, Amanda Askell, Anna Chen, Nova DasSarma, Dawn Drain, Stanislav Fort, Deep Ganguli, Tom Henighan, et~al. 2022.
\newblock Training a helpful and harmless assistant with reinforcement learning from human feedback.
\newblock \emph{arXiv preprint arXiv:2204.05862}.

\bibitem[{Bitchener and Basturkmen(2006)}]{bitchener2006perceptions}
John Bitchener and Helen Basturkmen. 2006.
\newblock Perceptions of the difficulties of postgraduate l2 thesis students writing the discussion section.
\newblock \emph{Journal of English for Academic Purposes}, 5(1):4--18.

\bibitem[{Blanchard et~al.(2013)Blanchard, Tetreault, Higgins, Cahill, and Chodorow}]{blanchard2013toefl11}
Daniel Blanchard, Joel Tetreault, Derrick Higgins, Aoife Cahill, and Martin Chodorow. 2013.
\newblock Toefl11: A corpus of non-native english.
\newblock \emph{ETS Research Report Series}, 2013(2):i--15.

\bibitem[{Bubeck et~al.(2023)Bubeck, Chandrasekaran, Eldan, Gehrke, Horvitz, Kamar, Lee, Lee, Li, Lundberg et~al.}]{bubeck2023sparks}
S{\'e}bastien Bubeck, Varun Chandrasekaran, Ronen Eldan, Johannes Gehrke, Eric Horvitz, Ece Kamar, Peter Lee, Yin~Tat Lee, Yuanzhi Li, Scott Lundberg, et~al. 2023.
\newblock Sparks of artificial general intelligence: Early experiments with gpt-4.
\newblock \emph{arXiv preprint arXiv:2303.12712}.

\bibitem[{Cao et~al.(2023)Cao, Yang, and Ng}]{cao2023mitigating}
Hannan Cao, Wenmian Yang, and Hwee~Tou Ng. 2023.
\newblock Mitigating exposure bias in grammatical error correction with data augmentation and reweighting.
\newblock In \emph{Proceedings of the 17th Conference of the European Chapter of the Association for Computational Linguistics}, pages 2123--2135.

\bibitem[{Cohen(1960)}]{cohen1960coefficient}
Jacob Cohen. 1960.
\newblock A coefficient of agreement for nominal scales.
\newblock \emph{Educational and psychological measurement}, 20(1):37--46.

\bibitem[{Cooley and Lewkowicz(1995)}]{cooley1995writing}
Linda Cooley and Jo~Lewkowicz. 1995.
\newblock The writing needs of postgraduate students at the university of hong kong: A project report.
\newblock \emph{Hong Kong papers in Linguistics and language teaching}, 18:121--123.

\bibitem[{Crossley et~al.(2016)Crossley, Kyle, and McNamara}]{crossley2016tool}
Scott~A Crossley, Kristopher Kyle, and Danielle~S McNamara. 2016.
\newblock The tool for the automatic analysis of text cohesion (taaco): Automatic assessment of local, global, and text cohesion.
\newblock \emph{Behavior research methods}, 48:1227--1237.

\bibitem[{Devlin et~al.(2018)Devlin, Chang, Lee, and Toutanova}]{devlin2018bert}
Jacob Devlin, Ming-Wei Chang, Kenton Lee, and Kristina Toutanova. 2018.
\newblock Bert: Pre-training of deep bidirectional transformers for language understanding.
\newblock \emph{arXiv preprint arXiv:1810.04805}.

\bibitem[{Dubois et~al.(2024)Dubois, Li, Taori, Zhang, Gulrajani, Ba, Guestrin, Liang, and Hashimoto}]{dubois2024alpacafarm}
Yann Dubois, Chen~Xuechen Li, Rohan Taori, Tianyi Zhang, Ishaan Gulrajani, Jimmy Ba, Carlos Guestrin, Percy~S Liang, and Tatsunori~B Hashimoto. 2024.
\newblock Alpacafarm: A simulation framework for methods that learn from human feedback.
\newblock \emph{Advances in Neural Information Processing Systems}, 36.

\bibitem[{Firdaus et~al.(2023)Firdaus, Wibawa, and Rahman}]{firdaus2023utilization}
Muhammad~Fatihul Firdaus, Joseph~Nugraha Wibawa, and Fajri~Fathur Rahman. 2023.
\newblock Utilization of gpt-4 to improve education quality through personalized learning for generation z in indonesia.
\newblock \emph{IT for Society}, 8(1).

\bibitem[{Gonz{\'a}lez(2017)}]{gonzalez2017contribution}
Melanie~C Gonz{\'a}lez. 2017.
\newblock The contribution of lexical diversity to college-level writing.
\newblock \emph{TESOL Journal}, 8(4):899--919.

\bibitem[{Halliday and Hasan(1976)}]{halliday2014cohesion}
Michael Alexander~Kirkwood Halliday and Ruqaiya Hasan. 1976.
\newblock \emph{Cohesion in english}.
\newblock Routledge.

\bibitem[{He et~al.(2021)He, Gao, and Chen}]{he2021debertav3}
Pengcheng He, Jianfeng Gao, and Weizhu Chen. 2021.
\newblock Debertav3: Improving deberta using electra-style pre-training with gradient-disentangled embedding sharing.
\newblock \emph{arXiv preprint arXiv:2111.09543}.

\bibitem[{Johnson et~al.(2016)Johnson, Acevedo, and Mercado}]{johnson2016vocabulary}
Mark~D Johnson, Anthony Acevedo, and Leonardo Mercado. 2016.
\newblock Vocabulary knowledge and vocabulary use in second language writing.
\newblock \emph{TESOL Journal}, 7(3):700--715.

\bibitem[{Knoch(2007)}]{knoch2007little}
Ute Knoch. 2007.
\newblock ‘little coherence, considerable strain for reader’: A comparison between two rating scales for the assessment of coherence.
\newblock \emph{Assessing writing}, 12(2):108--128.

\bibitem[{Lai and Tetreault(2018)}]{lai2018discourse}
Alice Lai and Joel Tetreault. 2018.
\newblock Discourse coherence in the wild: A dataset, evaluation and methods.
\newblock \emph{arXiv preprint arXiv:1805.04993}.

\bibitem[{Lautamatti(1978)}]{lautamatti1978observations}
Liisa Lautamatti. 1978.
\newblock Observations on the development of the topic in simplified discourse.
\newblock \emph{AFinLAn vuosikirja}, pages 71--104.

\bibitem[{Lorenz(1999)}]{lorenz1999learning}
Gunter Lorenz. 1999.
\newblock Learning to cohere: Causal links in native vs. non-native argumentative writing.
\newblock \emph{Pragmatics and Beyond New Series}, pages 55--76.

\bibitem[{Maimon and Tsarfaty(2023)}]{maimon2023cohesentia}
Aviya Maimon and Reut Tsarfaty. 2023.
\newblock Cohesentia: A novel benchmark of incremental versus holistic assessment of coherence in generated texts.
\newblock \emph{arXiv preprint arXiv:2310.16329}.

\bibitem[{McNamara et~al.(2010)McNamara, Louwerse, McCarthy, and Graesser}]{mcnamara2010coh}
Danielle~S McNamara, Max~M Louwerse, Philip~M McCarthy, and Arthur~C Graesser. 2010.
\newblock Coh-metrix: Capturing linguistic features of cohesion.
\newblock \emph{Discourse Processes}, 47(4):292--330.

\bibitem[{Naismith et~al.(2023)Naismith, Mulcaire, and Burstein}]{naismith2023automated}
Ben Naismith, Phoebe Mulcaire, and Jill Burstein. 2023.
\newblock Automated evaluation of written discourse coherence using gpt-4.
\newblock In \emph{Proceedings of the 18th Workshop on Innovative Use of NLP for Building Educational Applications (BEA 2023)}, pages 394--403.

\bibitem[{Omelianchuk et~al.(2020)Omelianchuk, Atrasevych, Chernodub, and Skurzhanskyi}]{omelianchuk2020gector}
Kostiantyn Omelianchuk, Vitaliy Atrasevych, Artem Chernodub, and Oleksandr Skurzhanskyi. 2020.
\newblock Gector--grammatical error correction: tag, not rewrite.
\newblock \emph{arXiv preprint arXiv:2005.12592}.

\bibitem[{OpenAI(2023)}]{openai_gpt4}
OpenAI. 2023.
\newblock Gpt-4 technical report.

\bibitem[{Ouyang et~al.(2022)Ouyang, Wu, Jiang, Almeida, Wainwright, Mishkin, Zhang, Agarwal, Slama, Ray et~al.}]{ouyang2022training}
Long Ouyang, Jeffrey Wu, Xu~Jiang, Diogo Almeida, Carroll Wainwright, Pamela Mishkin, Chong Zhang, Sandhini Agarwal, Katarina Slama, Alex Ray, et~al. 2022.
\newblock Training language models to follow instructions with human feedback.
\newblock \emph{Advances in neural information processing systems}, 35:27730--27744.

\bibitem[{Reinhart(1980)}]{reinhart1980conditions}
Tanya Reinhart. 1980.
\newblock Conditions for text coherence.
\newblock \emph{Poetics today}, 1(4):161--180.

\bibitem[{Schneider and Connor(1990)}]{schneider1990analyzing}
Melanie Schneider and Ulla Connor. 1990.
\newblock Analyzing topical structure in esl essays: Not all topics are equal.
\newblock \emph{Studies in second language acquisition}, 12(4):411--427.

\bibitem[{Tarnavskyi et~al.(2022)Tarnavskyi, Chernodub, and Omelianchuk}]{tarnavskyi2022ensembling}
Maksym Tarnavskyi, Artem Chernodub, and Kostiantyn Omelianchuk. 2022.
\newblock Ensembling and knowledge distilling of large sequence taggers for grammatical error correction.
\newblock \emph{arXiv preprint arXiv:2203.13064}.

\bibitem[{Touvron et~al.(2023)Touvron, Martin, Stone, Albert, Almahairi, Babaei, Bashlykov, Batra, Bhargava, Bhosale et~al.}]{touvron2023llama}
Hugo Touvron, Louis Martin, Kevin Stone, Peter Albert, Amjad Almahairi, Yasmine Babaei, Nikolay Bashlykov, Soumya Batra, Prajjwal Bhargava, Shruti Bhosale, et~al. 2023.
\newblock Llama 2: Open foundation and fine-tuned chat models.
\newblock \emph{arXiv preprint arXiv:2307.09288}.

\bibitem[{Yasunaga et~al.(2021)Yasunaga, Leskovec, and Liang}]{yasunaga2021lm}
Michihiro Yasunaga, Jure Leskovec, and Percy Liang. 2021.
\newblock Lm-critic: Language models for unsupervised grammatical error correction.
\newblock \emph{arXiv preprint arXiv:2109.06822}.

\bibitem[{Zhang et~al.(2024)Zhang, Chen, and Yu}]{zhang2024prolex}
Xuanming Zhang, Zixun Chen, and Zhou Yu. 2024.
\newblock Prolex: A benchmark for language proficiency-oriented lexical substitution.
\newblock \emph{arXiv preprint arXiv:2401.11356}.

\bibitem[{Zheng et~al.(2023)Zheng, Chiang, Sheng, Zhuang, Wu, Zhuang, Lin, Li, Li, Xing, Zhang, Gonzalez, and Stoica}]{zheng2023judging}
Lianmin Zheng, Wei-Lin Chiang, Ying Sheng, Siyuan Zhuang, Zhanghao Wu, Yonghao Zhuang, Zi~Lin, Zhuohan Li, Dacheng Li, Eric.~P Xing, Hao Zhang, Joseph~E. Gonzalez, and Ion Stoica. 2023.
\newblock \href {https://arxiv.org/abs/2306.05685} {Judging llm-as-a-judge with mt-bench and chatbot arena}.
\newblock \emph{Preprint}, arXiv:2306.05685.

\bibitem[{Zhou et~al.(2024)Zhou, Liu, Xu, Iyer, Sun, Mao, Ma, Efrat, Yu, Yu et~al.}]{zhou2024lima}
Chunting Zhou, Pengfei Liu, Puxin Xu, Srinivasan Iyer, Jiao Sun, Yuning Mao, Xuezhe Ma, Avia Efrat, Ping Yu, Lili Yu, et~al. 2024.
\newblock Lima: Less is more for alignment.
\newblock \emph{Advances in Neural Information Processing Systems}, 36.

\end{thebibliography}

\appendix

\clearpage

\section{Detailed annotation scheme for \textit{DECOR}}
\label{app:annotation_scheme}

\subsection{Incoherence detection and reasoning}
In the coherence detection process, coherent $(C, S)$ pairs are marked with a $1$, while incoherent ones are marked with a $-1$. For cases unrelated to writing coherence (e.g., sentence parsing errors), a $0$ is assigned and they will be excluded from the resulting dataset.

To complete this task, annotators were instructed that for each sentence there is a topic $T$, and a context $C$, which comprises all preceding sentences up to and immediately before sentence $S$ in the essay and for all incoherent sentences to provide all possible reasons (R1-R7) for the break in coherence. They were also instructed to determine if each sentence $S$ is coherent with context $C$ based on the provided instructions. Lastly, for each incoherent sentence $S$, annotators were asked to revise $S$ to improve its coherence, taking into account the types of edits suggested for each identified reason.
Below is the complete list of reasons that were provided to the annotators.
\begin{itemize}
\item (1) The sentence $S$ is coherent with the context $C$ as:
\begin{itemize}
\item The sentence $S$ semantically connects to the context $C$, (i.e. with proper use of reference words, repeated words/ ideas, and substitution), and
\item All entities discussed in the new sentence $S$ have been introduced in $C$, and
\item The new sentence $S$ demonstrates reasonable discourse relation with previous ones, and
\item The new sentence $S$ contains a meaning consistent with previously presented data in $C$ and 
\item The new sentence $S$ contains a meaning relevant to previously presented data in $C$
\end{itemize}

\item (-1) The sentence $S$ is not coherent with $C$ as: 
\begin{itemize}
\item R1: (Semantic connection) The sentence $S$ does not connect semantically with the context $C$;
\item R2: (Entity reference) The new sentence $S$ discusses an entity that has not been introduced in $C$ yet, or the new sentence $S$ discusses an entity that is ambiguous in $C$ or
\item R3: (Discourse relation) The relation between sentence $S$ and previous ones in $C$ doesn't make sense due to a missing discourse marker.
\item R4: (Consistency) The new sentence $S$ contradicts or is inconsistent with previously presented information, or 
\item R5: (Contextual relevance) The new sentence $S$ introduces information that is completely irrelevant to the context 
\item R6: (Tangential relevance) The new sentence $S$ introduces information that is either tangential or slightly irrelevant to the context.
\item R7: (Others) Other reasons that are not listed above. For example, the comment (rheme/focus) of the sentence does not agree with the topic of the sentence.
\end{itemize}
\item (0) Other cases that have nothing to do with writing coherence
\end{itemize}

For incoherent reasons, annotators were asked to mark “1” in the corresponding reason column of the annotation document and leave the others empty. For example, if sentence $S$ is incoherent to context $C$ due to reason 2 (Entity reference) and reason 3 (Discourse relation), mark “1” in both R2 and R3 columns, and leave the others empty.

\subsection{Types of edits for incoherent sentence rewriting}
\label{app:edits}
Given an incoherent sentence-context pair $(C, S)$, annotators are instructed to make the least invasive changes to rewrite sentence $S$. The suggested edits are described as follows:
\begin{itemize}
    \item \emph{Semantic connection}: add reference words or repeated words/ideas or substitution that can semantically connect sentence $S$ to context $C$.
    \item \emph{Entity reference}: link the newly introduced entity or ambiguous entity in sentence $S$ to context $C$.
    \item \emph{Discourse relation}: add or change a discourse marker that ties sentence $S$ with context $C$.
    \item \emph{Consistency}: align the newly introduced information in sentence $S$ with previously introduced information in context $C$ so that the new information does not contradict the context.
    \item \emph{Contextual relevance}: modify sentence $S$ so that it is relevant to the context established by the writer.
    \item \emph{Tangential relevance}: delete the sentence and edit with "DELETE".
    \item \emph{Others}: rewrite the sentence so that the comment of the sentence agrees with the topic of the sentence.
\end{itemize}

Note that we recommend "DELETE" if sentence $S$ is tangential, as its presence following context $C$ is unnecessary.

\section{More details about inter-annotator agreement}
\label{app:iaa_reason_type}
\subsection{Inter-annotator agreement scores across different reason types}

The specific inter-annotator agreement scores for both incoherence detection and reasoning tasks are shown in Table \ref{tab:iaa_scores}. Overall, our annotators achieved very high agreement on both tasks.

% \begin{table}[]
% \centering
% \begin{tabular}{lc}
% \toprule
% Group       & Cohen's Kappa \\ \midrule
% Coherence   & 0.83          \\ \midrule
% Cohesion    & 0.80          \\
% Consistency & 1.00             \\
% Relevance   & 0.86          \\
% Others      & 1.00             \\ \bottomrule
% \end{tabular}
% \end{table}

% \begin{table}[]
% \centering
% \begin{tabular}{lc}
% \toprule
% Label Code & Cohen's Kappa \\ \midrule
% R1         & 0.84          \\
% R2         & 0.74          \\
% R3         & 0.88          \\
% R4         & 1.00          \\
% R5         & 1.00          \\
% R6         & 0.86          \\
% R7         & 1.00          \\ \bottomrule
% \end{tabular}
% \end{table}

\begin{table}[ht]
    \centering
    \begin{subtable}[t]{.8\columnwidth}
    \centering
    \begin{adjustbox}{max width=.75\linewidth}
        \centering
        \begin{tabular}{rl}
\toprule
Group       & Cohen's $\kappa$ \\ \midrule
Coherence   & 0.83          \\ \midrule
Cohesion    & 0.80          \\
Consistency & 1.00             \\
Relevance   & 0.86          \\
Others      & 1.00             \\ \bottomrule
\end{tabular}
\end{adjustbox}
        \caption{Inter-annotator agreement on incoherence detection and reasons clustered into groups.}
    \end{subtable}
    \\
    \vspace{10pt}
    \begin{subtable}[t]{.8\columnwidth}
    \centering
    \begin{adjustbox}{max width=.75\linewidth}
        \centering
        \begin{tabular}{rl}
\toprule
Reasons & Cohen's $\kappa$ \\ \midrule
R1         & 0.84          \\
R2         & 0.74          \\
R3         & 0.88          \\
R4         & 1.00          \\
R5         & 1.00          \\
R6         & 0.86          \\
R7         & 1.00          \\ \bottomrule
\end{tabular}
\end{adjustbox}
        \caption{Inter-annotator agreement on specific reasons.}
    \end{subtable} \\
\caption{Inter-annotator agreement scores for annotations.}
\label{tab:iaa_scores}
\end{table}

% \begin{figure}[t]
%   \includegraphics[width=\columnwidth]{example-image-golden}
%   \caption{IAA scores across different reason types.}
%   % \label{tab:sent_word_essay}
% \end{figure}

\subsection{The achievability of high inter-annotator agreement on \textit{DECOR}}

We achieve a high inter-rater agreement for our dataset through our meticulously structured and clearly defined annotation process and scheme. Specifically, we recruited two expert annotators who are both professors with extensive experience in teaching English as a foreign language and have advanced degrees in Applied Linguistics. Before starting, we delivered a comprehensive tutorial and thoroughly reviewed our detailed annotation guidelines with the annotators. To familiarize them with our annotation scheme, we selected eight medium-level essays from TOEFL-11, generating 109 context-sentence pairs. The tutorial was structured into three sessions to ensure thorough coverage of all instances. During these sessions, annotators collaboratively labeled the samples. We also refined the annotation scheme as necessary. Additionally, we incorporated examples of errors encountered during these tutorial sessions into the annotation scheme. Subsequently, to quantitatively assess the inter-annotator agreement, we sampled an additional five medium-level essays, distinct from those used in the tutorial sessions, resulting in 72 context-sentence pairs.

Among the 72 samples, the first annotator labeled 35 context-sentence pairs as incoherent, and 30 as coherent. The second annotator labeled 39 pairs as incoherent, and 28 as coherent. Note that the rest were labeled as “uncertain”. Therefore, the resulting annotations for the 72 samples were relatively balanced. Given the high inter-annotator agreement achieved by the two annotators, with Cohen’s Kappa scores of 0.83 for incoherence detection and 0.90 for incoherence reasoning, we subsequently assigned them to independently annotate the entire test set.

\section{Additional statistics of \textit{DECOR}}
\begin{figure}[t]
\centering
  \includegraphics[width=0.35\textwidth]{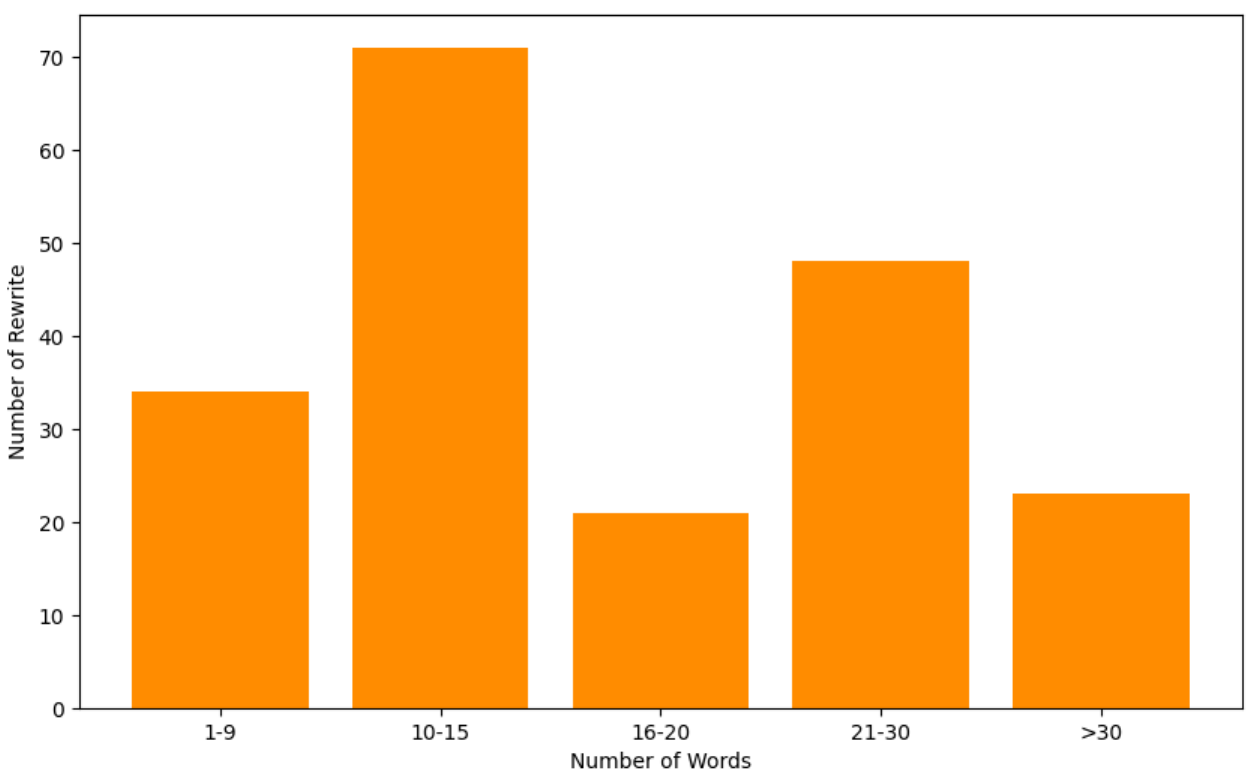}
  \caption{The number of words per rewrite.}
  \label{fig:words_rewrite}
  % {tab:words_rewrite}
\end{figure}

\begin{figure}[htbp]
    \centering
    \begin{subfigure}{0.23\textwidth}
        \centering
        \includegraphics[width=\textwidth]{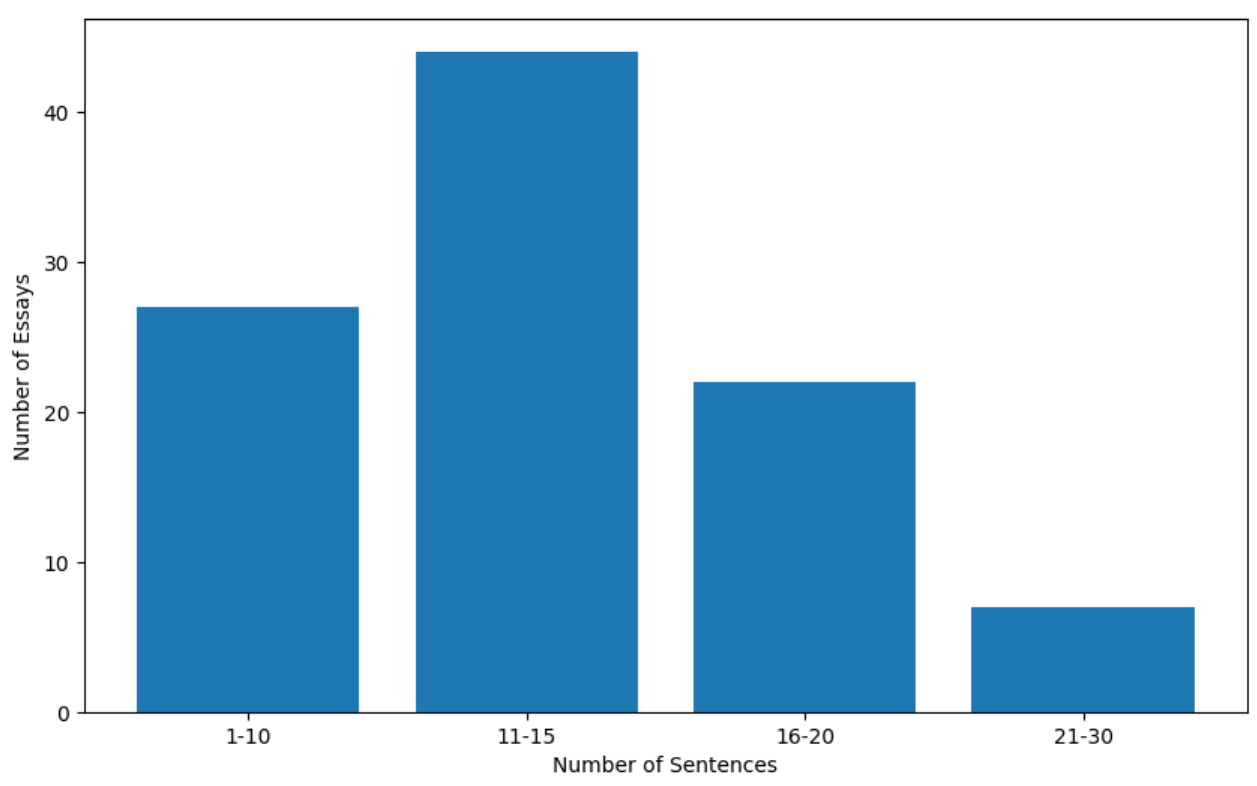}
        \caption{The number of sentences per essay.}
        \label{fig:sub1}
    \end{subfigure}\hfill
    \begin{subfigure}{0.23\textwidth}
        \centering
        \includegraphics[width=\textwidth]{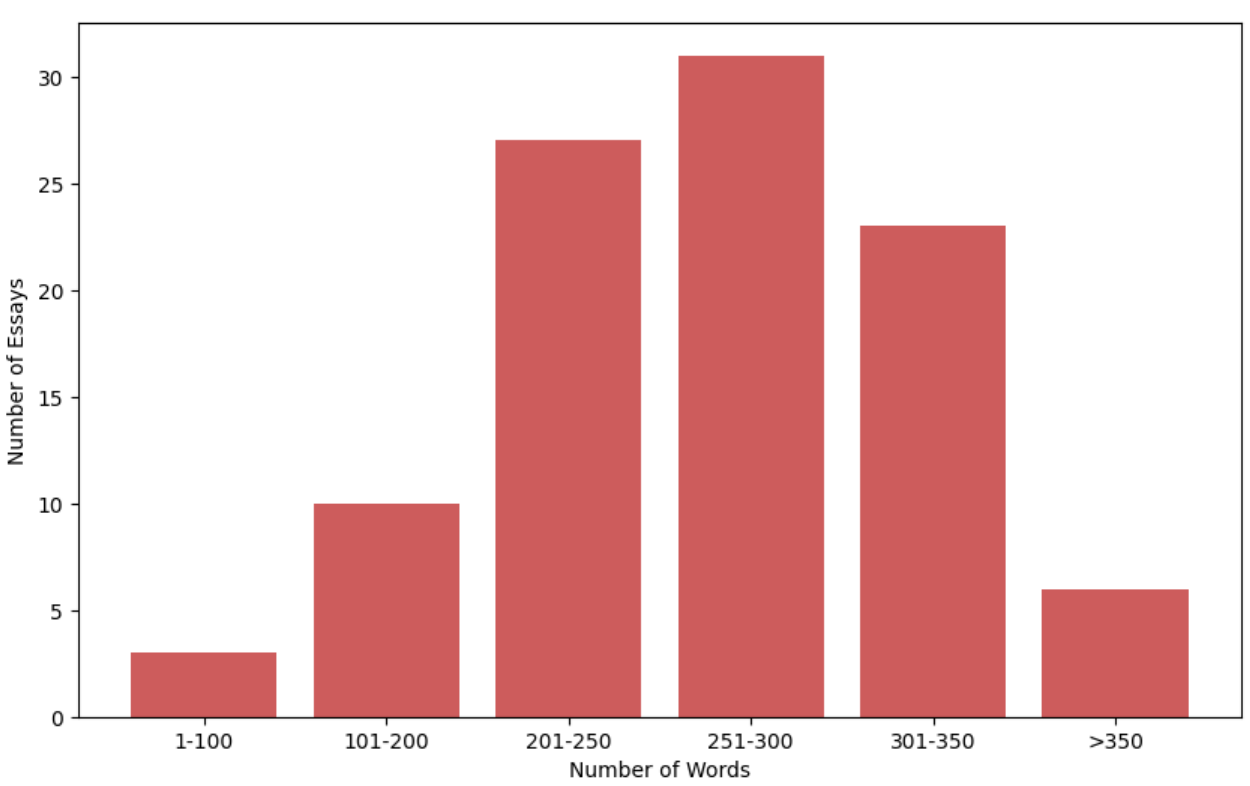}
        \caption{The number of words per essay.}
        \label{fig:sub2}
    \end{subfigure}
    \caption{Distribution of essays by number of sentences and number of words.}
\label{fig:other_stats}
\end{figure}

We show the overall distribution of rewrite lengths measured by the number of words in Figure \ref{fig:words_rewrite}. We also illustrate the distribution of essays measured by the number of sentences and words in Figure \ref{fig:other_stats}.

\section{Human evaluation details}
\label{app:human_eval_d}
During human evaluation, our human expert was asked to compare system-generated rewrites with those produced by GPT-4. Specifically, we conducted three sets of pair-wise comparisons: 1) human-generated rewrites VS GPT-4; 2) rewrites generated by Llama2-7B trained with reasons VS GPT-4; and 3) rewrites generated by Llama2-7B trained without reasons VS GPT-4. For each set of pair-wise comparisons, for each pair of outputs (e.g. system 1 output VS system 2 output) the human expert was asked “Between system 1 and system 2 outputs, select the output that provides better rewrite based on the reason for incoherence; otherwise, choose “Tie” to indicate equal quality”.

After human evaluation, we interviewed the human expert and asked for his feedback on the system outputs. In general, the evaluator was surprised that our fine-tuned system can generate rewrites of decent quality. Among the three sets of comparisons, he was not able to tell which one was generated by humans. This indicates that our fine-tuned smaller language models are already capable of correcting the incoherence well based on the reasons behind it.

\section{Synthesizing training data with GPT-4}
\subsection{Dataset Synthesis}
\label{app:syn_prompt_gpt}

A large portion of the training set of \textit{DECOR} is synthesized by GPT-4 based on human-annotated examples in order to increase generalizability and variety. Table~\ref{tab:gpt_syn_prompt} shows the prompt we used.

\begin{table*}[]
\centering
\begin{adjustbox}{max width=\textwidth}
\begin{tabular}{l}
\hline
\begin{tabular}[c]{@{}l@{}}
You are an English teacher aiming to improve coherence in student writing. You are about to synthesize data for the coherence detection task. Concretely, for each data point, \\you will be given: a sentence S and a context C, which comprises all preceding sentences up to and immediately before sentence S in an essay written by an English second \\language learner. Then, you should follow the following steps to create a complete data point:\\
1) For sentence S and context C, determine if sentence S is coherent with context C. You need to output 1 for [Coherence] if the sentence S is coherent when appended to the \\context C; otherwise, output 0;\\
2) Then, if you output 1 in the previous step, output "Done" and finish; otherwise, move on to the following steps;\\
3) You need to output 1 for [Reason 1] if the sentence S does not connect semantically with the context C; otherwise, output 0;\\
4) You need to output 1 for [Reason 2] if the new sentence S discusses an entity that has not been introduced in C yet, or the new sentence S discusses an entity that is \\ambiguous in C; otherwise, output 0;\\
5) You need to output 1 for [Reason 3] if the relation between sentence S and previous ones in C doesn’t make sense due to a missing discourse marker; otherwise, output 0;\\
6) You need to output 1 for [Reason 4] if the new sentence S contradicts or is inconsistent with previously presented information in C; otherwise, output 0;\\
7) You need to output 1 for [Reason 5] if the new sentence S introduces information that is completely irrelevant to the context C; otherwise, output 0;\\
8) You need to output 1 for [Reason 6] if the new sentence S introduces information that is either tangential or slightly irrelevant to the context C; otherwise, output 0;\\
9) You need to output 1 for [Reason 7] if the comment (rheme/focus) of the sentence does not agree with the topic of the sentence; otherwise, output 0\\
10) [Rewrite] You should modify sentence S as minimally as possible to improve its coherence based on the following suggestions for each reason you might select above:\\
- [Reason 1]: add reference words or repeated words or substitutions that can semantically connect sentence S to the context C;\\
- [Reason 2]: link the newly introduced entity or ambiguous entity in S to the given context C\\
- [Reason 3]: add or change a discourse marker that ties the sentence S with the given context C\\
- [Reason 4]: align the newly introduced information with previously introduced information so that the new information in S does not contradict the context C\\
- [Reason 5]: modify the sentence S so that it is relevant to the context C established by the writer\\
- [Reason 6]: only output "DELETE" for deleting the sentence S\\
- [Reason 7]: rewrite sentence S so that the comment of sentence S agrees with the topic of sentence S\\
Please disregard any incoherences in context C. You should output 1 for [Coherence] only if:\\
a) sentence S semantically connects to context C, and\\
b) all entities discussed in the new sentence S have been introduced in C, and\\
c) sentence S demonstrates reasonable discourse relation with previous ones, and\\
d) sentence S contains a meaning consistent with previously presented data in C, and\\
e) sentence S contains a meaning relevant to previously presented data in C.\\
Here are some examples:\\
C: I believe that young people nowadays do not give enough time to helping their communities.\\
S: This, i believe is caused by the environment we live in.\\
- [Coherence]: 1\\
- Done\\
C: Then, I wanna indicate that young people can study many things that are interesting or exciting things for young people.\\
S: About students, they can learn various fields that students want to study.\\
- [Coherence]: 0\\
- [Reason 1]: 1\\
- [Reason 2]: 0\\
- [Reason 3]: 1\\
- [Reason 4]: 0\\
- [Reason 5]: 0\\
- [Reason 6]: 0\\
- [Reason 7]: 0\\
- [Rewrite]: For example when they study, they can learn various fields that they want to study.\\
C: There are three main reasons that my ideas support effectively, like action, study and knowledge.\\
S: First of all, I wanna introduce young people's active points in comparison with older people.\\
- [Coherence]: 0\\
- [Reason 1]: 0\\
- [Reason 2]: 0\\
- [Reason 3]: 0\\
- [Reason 4]: 1\\
- [Reason 5]: 0\\
- [Reason 6]: 0\\
- [Reason 7]: 0\\
- [Rewrite]: First of all, I wanna introduce young people's actions in comparison with older people's.\\
C: These publicity agents use a lot of techniques to make the products look better, for example they use specialized software like photoshop to increase the size of the \\product or make it brighter, or maybe an artificial imitation of the product that does not necessarily have the same texture of look.\\
S: Even though one can observe this situation mostly in food products.\\
- [Coherence]: 0\\
- [Reason 1]: 0\\
- [Reason 2]: 0\\
- [Reason 3]: 0\\
- [Reason 4]: 0\\
- [Reason 5]: 0\\
- [Reason 6]: 1\\
- [Reason 7]: 0\\
- [Rewrite]: DELETE\\
C: I, however, think in terms of physical and mental factors young people are superior to older people.\\
S: For example, in the case of sports young people can run and jump, and they can train their muscles that are used in each sport such as transitional sports or silence sports.\\
- [Coherence]: 0\\
- [Reason 1]: 0\\
- [Reason 2]: 0\\
- [Reason 3]: 0\\
- [Reason 4]: 0\\
- [Reason 5]: 0\\
- [Reason 6]: 0\\
- [Reason 7]: 1\\
- [Rewrite]: For example, in the case of sports young people can run and jump, and they can train their muscles for sports more than older people can.\\
Now, please generate:
\end{tabular} \\ \hline
\end{tabular}
\end{adjustbox}
\caption{The prompt for GPT-4 to generate synthetic training data. We provide 4 human-annotated examples in order to constrain its output format.}
\label{tab:gpt_syn_prompt}
\end{table*}

\subsection{Post-Processing}
\label{app:syn_data_postprocessing}
As described in Section \ref{sec:task2_data}, we prompted GPT-4 to identify all potential reasons for each incoherent context-sentence pair. To obtain the training data for detecting \textit{Consistency} as the cause, an instance is labeled as "Yes" if GPT-4 identifies R4 as the cause of incoherence for that instance; otherwise, the label is "No". We conducted similar post-processing steps to create the training data for \textit{Relevance} and \textit{Others} tasks. Given that the initial data after post-processing is extremely unbalanced for each sub-task. We downsampled instances of the majority class to achieve a more balanced training dataset. The statistics of the resulting data for each sub-task are shown in Table \ref{tab:post-proc-stats}.

\begin{table}[t]
\centering
\begin{adjustbox}{max width=.96\linewidth}
\begin{tabular}{lrrrr}
\toprule
\multicolumn{1}{c}{Label} & \multicolumn{1}{c}{Cohesion} & \multicolumn{1}{c}{Consistency} & \multicolumn{1}{c}{Relevance} & \multicolumn{1}{c}{Other} \\ \midrule
Yes                       & 827                          & 511                             & 460                           & 387                       \\
No                        & 825                          & 848                             & 848                           & 848                       \\ \bottomrule
\end{tabular}
\end{adjustbox}
\caption{Statistics of synthetic training data for the incoherence reasoning task.}
\label{tab:post-proc-stats}
\end{table}

% \todo{discuss the post-processing steps to create training data for each of the four sub-tasks.}
% \todo{stats of the training data for each sub-task}

\section{Details of Experiments}
\label{app:details_baseline}

\subsection{Classification-based models}

\begin{table*}[t]
\centering
\begin{adjustbox}{max width=.7\textwidth}
\begin{tabular}{rl|ccccc}
\toprule
\multicolumn{1}{c|}{\multirow{2}{*}{\textbf{Models}}} & \multicolumn{1}{c|}{\multirow{2}{*}{\textbf{\begin{tabular}[c]{@{}c@{}}Training \\ Data \end{tabular}}}} & \multicolumn{1}{c|}{\multirow{2}{*}{\textbf{\begin{tabular}[c]{@{}c@{}}Incoherence \\ Detection (\%)\end{tabular}}}} & \multicolumn{4}{c}{\textbf{Incoherence Reasoning Selection (\%)}} \\ \cmidrule{4-7} 
\multicolumn{1}{c|}{} & \multicolumn{1}{c|}{} & \multicolumn{1}{c|}{} & \textit{Cohesion} & \textit{Consistency} & \textit{Relevance} & \textit{Others}     \\ \midrule
\multicolumn{2}{l|}{\textit{\textbf{Classification-based}}} & \multicolumn{1}{l}{} & \multicolumn{1}{l}{} & \multicolumn{1}{l}{} & \multicolumn{1}{l}{} & \multicolumn{1}{l}{} \\
\multirow{3}{*}{BERT-base}    & $D_C$       & 63.04 & 48.17 & 93.76 & 28.47 & - \\
                              & $D_T$       & 66.43 & 44.38 & 75.41 & 46.37 & 80.36 \\ 
                              & $D_C + D_T$ & 65.64 & 47.64 & 80.40 & 46.52 & - \\ \cmidrule{2-7}
\multirow{3}{*}{BERT-large}   & $D_C$       & 64.21 & 45.93 & 93.76 & 28.47 & - \\
                              & $D_T$       & 65.71 & 44.75 & \textbf{93.99} & \textbf{48.34} & 82.65  \\ 
                              & $D_C + D_T$ & 66.26 & 45.67 & \textbf{93.99} & 42.00 & - \\ \cmidrule{2-7} 
\multirow{3}{*}{DeBERTa-base} & $D_C$       & 62.21 & 47.93 & 93.88 & 29.45 & - \\
                              & $D_T$       & \textbf{68.54} & 46.47 & 77.17 & 45.14 & 74.20 \\ 
                              & $D_C + D_T$ & 67.52 & \textbf{48.50} & 77.97 & 45.53 & - \\ \cmidrule{2-7} 
\multirow{3}{*}{DeBERTa-large}& $D_C$       & 53.78 & 45.93 & \textbf{93.99} & 28.47 & - \\
                              & $D_T$       & 53.78 & 45.93 & 92.74 & 41.36 & \textbf{89.70} \\ 
                              & $D_C + D_T$ & 67.05 & 47.68 & 84.73 & 47.02 & - \\ \midrule
\multicolumn{2}{l|}{\textbf{\textit{Generation-based}}} & \multicolumn{1}{l}{} & \multicolumn{1}{l}{} & \multicolumn{1}{l}{} & \multicolumn{1}{l}{} & \multicolumn{1}{l}{} \\
\multirow{3}{*}{Llama2-7B} & $D_C$       & 59.52 & 43.93 & 93.65 & 28.87 & - \\
                           & $D_T$       & 66.08 & 46.63 & 83.55 & 47.20 & 87.78 \\ 
                           & $D_C + D_T$ & 67.29 & 43.40 & 88.26 & 45.65 & - \\ \bottomrule
\end{tabular}
\end{adjustbox}
\caption{Evaluation of more BERT and DeBERTa models on DECOR using weighted F1 scores for Incoherence Detection and Incoherence Reasoning tasks. Task-specific synthetic data that was used in training is denoted as $D_T$ and out-of-distribution training data from \citet{maimon2023cohesentia} is denoted as $D_C$. For each task, we also combine their respective $D_C$ and $D_T$ to train models.}
\label{tab:baseline-app}
\end{table*}
For training the BERT and DeBERTa models, we established our pipeline based on the platform developed by \cite{maimon2023cohesentia}. Specifically, for the incoherence detection and reasoning tasks, we train these two models on both the CoheSentia dataset $D_C$ and the synthetic training data $D_T$, as well as a combination of the two, $D_C + D_T$. For validation purposes, we utilized the existing validation dataset from CoheSentia. Additionally, we allocated 10\% of the synthetic dataset for evaluating models trained with $D_T$. Note that since \textit{DECOR} and CoheSentia has different definitions for the Others category, it is not possible to evaluate a model trained with $D_C$ for this category on \textit{DECOR}, nor does it make sense to combine the datasets for this category. All models are trained for 10 epochs on their respective dataset with a learning rate of $2 \times 10^{-5}$ and batch size of 8 on a single NVIDIA A100-80G GPU. Based on the results from the validation set, we evaluate the best checkpoint on \textit{DECOR} for each task.

\subsection{Generation-based models}
For the task of incoherence detection and reasoning, we fine-tuned Llama2-7B under three experiment settings (i.e. $D_T$, $D_C$, and $D_C + D_T$). For the task of incoherence rewriting, besides Llama2-7B, we additionally fine-tuned Llama3-8B-Instruct on our synthetic training data generated by GPT-4. Specifically, we referred to the platform developed by \citet{zheng2023judging} to construct our training pipeline. For all settings, we fine-tuned it for a maximum of 5 epochs, using a single NVIDIA A100-80G GPU. Additionally, we configured the training batch size per device to 1 and established the initial learning rate at $1 \times 10^{-5}$, with a linear learning rate scheduler. The best checkpoints were selected based on the performance on the validation data.

\subsection{Additional results}
\label{sec:extra_results}
The additional results for the incoherence detection and reasoning tasks are shown in Table~\ref{tab:baseline-app}. We can see that training with the DECOR training set $D_T$ usually outperforms training with the out-of-distribution dataset $D_C$, and training with the combined dataset $D_C + D_T$ can result in a performance uplift, possibly thanks to greater generalizability. A notable point is that for the consistency test set, the labels are relatively imbalanced as mentioned in Section~\ref{sec:ann_process_statistics}, i.e. there are a lot more consistent examples than inconsistent examples, so models that tend to bias towards predicting all examples as consistent would score higher at weighted F1. This could be corrected by using macro F1 or expanding the test set to include more inconsistent examples, which we plan to explore in the future.

\begin{table*}[t]
\centering
\begin{adjustbox}{max width=\linewidth}
\begin{tabular}{l|l|l|l|l|l}
\toprule
\multicolumn{1}{c|}{\textbf{Models}} & \multicolumn{1}{c|}{\textbf{Context}} & \multicolumn{1}{c|}{\textbf{Current Sentence}} & \multicolumn{1}{c|}{\textbf{Reasons}} & \multicolumn{1}{c|}{\textbf{Condition}} & \multicolumn{1}{c}{\textbf{Rewrite}} \\ \hline
\multirow{3}{*}{Llama2-7B} & \multirow{3}{*}{\begin{tabular}[c]{@{}l@{}}So many investors struggled to make extraordinary shops. \\ And they also have to care about the location of the shop \\ and the menu. \textcolor{blue}{Because there are rival shops}.\end{tabular}} & \multirow{3}{*}{\begin{tabular}[c]{@{}l@{}}\textcolor{red}{But} if someone tries a new thing, he \\ doesn't have to worry about the rival.\end{tabular}} & \multirow{3}{*}{\begin{tabular}[c]{@{}l@{}}Discourse\\ marker\end{tabular}} & w/ reason & \begin{tabular}[c]{@{}l@{}}\textbf{\textit{However}}, if someone tries a new thing, \\ he doesn't have to worry about the rival.\end{tabular} \\ \cmidrule{5-6}
 & & & & w/o reason & \begin{tabular}[c]{@{}l@{}}\textcolor{red}{For example}, if someone tries a new thing, \\ he doesn't have to worry about the rival.\end{tabular} \\ \midrule
\multirow{3}{*}{\begin{tabular}[c]{@{}l@{}}Llama3-8B\\ -Instruct \end{tabular}} & \multirow{3}{*}{\begin{tabular}[c]{@{}l@{}}I found in several books of scientists, universities magazines, \\ that people who \textcolor{blue}{want to succeed} need to take risks, risk in \\ research, risk in budgets, contracts, borrow investment.\end{tabular}} & \multirow{3}{*}{\begin{tabular}[c]{@{}l@{}}You can never \textcolor{red}{do big money} with \\the regular risks.\end{tabular}} & \multirow{3}{*}{\begin{tabular}[c]{@{}l@{}} Semantic\\ connection\end{tabular}}  & w/ reason & \begin{tabular}[c]{@{}l@{}}You can never \textbf{\textit{achieve big success}} without \\ taking risks.\end{tabular} \\ \cmidrule{5-6}
 & & & & w/o reason & \begin{tabular}[c]{@{}l@{}}\textcolor{red}{For example}, you can never \textcolor{red}{do big business} \\ with regular risks.\end{tabular} \\ \bottomrule
\end{tabular}
\end{adjustbox}
\caption{Example rewrites produced by our fine-tuned models, using incoherent context-sentence pairs as input. The reason for the incoherence is also specified. The parts of the sentence that is causing the incoherence are marked as \textcolor{red}{red}. Important information from the context is marked with \textcolor{blue}{blue}. These examples demonstrate that fine-tuning with reasons for incoherence yields better rewrites compared to those produced by the model trained without such reasons.}
\label{tab:qual_examples_task3}
\end{table*}

\section{GPT-4 Prompts for Detection, Reasoning, and Rewriting}
\label{app:gpt4_all_prompts}

To leverage the in-context learning capabilities of LLMs, we also prompt GPT-4 in a zero-shot and few (16)-shot setting to establish our baseline results.

\subsection{Detection}

For coherence detection, Table~\ref{tab:gpt_detection_prompt_zero} shows the zero-shot prompt while Table~\ref{tab:gpt_detection_prompt_16} shows the 16-shot prompt.

\begin{table*}[]
\centering
\begin{adjustbox}{max width=\textwidth}
\begin{tabular}{l}
\hline
\begin{tabular}[c]{@{}l@{}}
You are about to perform the task of coherence detection for the sentences written by second-language English learners. \\In this task, given a sentence S and a context C, you need to output 1 if S is coherent with C based on the following instructions; \\otherwise, output 0. You should output 1 only if: \\a) sentence S semantically connects to context C, and \\b) all entities discussed in the new sentence S have been introduced in C, and \\c) the relation between sentence S and previous ones in C makes sense due to proper use of discourse markers, and \\d) the new sentence S does not contradict or is not inconsistent with previously presented information in C, and \\e) the new sentence S introduces information that is relevant to the context C established by the writer. \\Now, please generate:\\\\
C: [context]\\
S: [sentence]
\end{tabular} \\ \hline
\end{tabular}
\end{adjustbox}
\caption{Zero-shot prompt for GPT-4 coherence detection.}
\label{tab:gpt_detection_prompt_zero}
\end{table*}

\begin{table*}[]
\centering
\begin{adjustbox}{max width=\textwidth}
\begin{tabular}{l}
\hline
\begin{tabular}[c]{@{}l@{}}
You are about to perform the task of coherence detection for the sentences written by second-language English learners. \\In this task, given a sentence S and a context C, you need to output 1 if S is coherent with C; otherwise, output 0 and provide \\a concise explanation. Please disregard any incoherences in context C. Specifically, output 0 if: \\
a) the sentence S does not connect semantically with the context C; or\\
b) the new sentence S discusses an entity that has not been introduced in C yet, or the new sentence discusses an entity that is \\ambiguous in C; or\\
c) the relation between sentence S and previous ones in C doesn't make sense due to an inaccurate discourse marker; or\\
d) sentence S contradicts or is inconsistent with previously presented information in C; or\\
e) sentence S introduces information that is completely irrelevant to the context C; or\\
f) sentence S introduces information that is either tangential or slightly irrelevant to the context C; or\\
g) the comment of the sentence does not agree with the topic of the sentence itself, or some terms in S are not semantically \\consistent with each other.\\

Here are some examples:\\
C: I believe that young people nowadays do not give enough time to helping their communities.\\
S: This, i believe is caused by the environment we live in.\\
Answer: 1\\

... (14 more examples)\\

C: I, however, think in terms of physical and mental factors young people are superior to older people.\\
S: For example, in the case of sports young people can run and jump, and they can train their muscles that are used in each sport \\such as transitional sports or silence sports.\\
Answer: 0\\
Concise explanation: "transitional sports" and "silence sports" are not semantically consistent with each other. They also do not \\agree with the topic of the sentence.\\

Now, please generate:\\\\
C: [context]\\
S: [sentence]\\
Answer:
\end{tabular} \\ \hline
\end{tabular}
\end{adjustbox}
\caption{16-shot prompt for GPT-4 coherence detection.}
\label{tab:gpt_detection_prompt_16}
\end{table*}

\subsection{Reasoning - Cohesion}

For reasoning about the current sentence's cohesion, Table~\ref{tab:gpt_cohesion_prompt_zero} shows the zero-shot prompt while Table~\ref{tab:gpt_cohesion_prompt_16} shows the 16-shot prompt.

\begin{table*}[]
\centering
\begin{adjustbox}{max width=\textwidth}
\begin{tabular}{l}
\hline
\begin{tabular}[c]{@{}l@{}}
In this task, given a sentence S that is incoherent with a context C, you need to detect the reason that causes the incoherence. \\There are seven possible reasons that can cause incoherences: \\a) sentence S is incoherent with C because S does not connect semantically with C; \\b) sentence S is incoherent with C because S discusses an entity that has not been introduced in C yet, or the new sentence S \\discusses an entity that is ambiguous in C; \\c) sentence S is incoherent with C because the discourse relation between S and previous ones in C doesn't make sense due to \\an incorrect discourse marker; \\d) sentence S is incoherent with C because S contradicts or is inconsistent with previously presented information in C; \\e) sentence S is incoherent with C because S introduces information that is completely irrelevant to the context C; \\f) sentence S is incoherent with C because S introduces information that is tangential and unnecessary; \\g) sentence S is incoherent with C because the comment of the sentence does not agree with the topic of the sentence\\ In this task, please think step by step and output 1 only if S is incoherent with C due to any of reason a), reason b) or reason c). \\Otherwise, output 0 if S is incoherent with C due to other reasons. In your answer, start by directly generating either 1 or 0, \\then followed with reasons. Now, please generate the answer:\\\\
C: [context]\\
S: [sentence]
\end{tabular} \\ \hline
\end{tabular}
\end{adjustbox}
\caption{Zero-shot prompt for GPT-4 cohesion reasoning.}
\label{tab:gpt_cohesion_prompt_zero}
\end{table*}

\begin{table*}[]
\centering
\begin{adjustbox}{max width=\textwidth}
\begin{tabular}{l}
\hline
\begin{tabular}[c]{@{}l@{}}
In this task, given a sentence S that is incoherent with a context C, you need to detect the reason that causes the incoherence. \\There are seven possible reasons that can cause incoherences: \\a) sentence S is incoherent with C because S does not connect semantically with C; \\b) sentence S is incoherent with C because S discusses an entity that has not been introduced in C yet, or the new sentence S \\discusses an entity that is ambiguous in C; \\c) sentence S is incoherent with C because the discourse relation between S and previous ones in C doesn't make sense due to \\an incorrect discourse marker; \\d) sentence S is incoherent with C because S contradicts or is inconsistent with previously presented information in C; \\e) sentence S is incoherent with C because S introduces information that is completely irrelevant to the context C; \\f) sentence S is incoherent with C because S introduces information that is tangential and unnecessary; \\g) sentence S is incoherent with C because the comment of the sentence does not agree with the topic of the sentence\\ In this task, please think step by step and output 1 only if S is incoherent with C due to any of reason a), reason b) or reason c). \\Otherwise, output 0 if S is incoherent with C due to other reasons. In your answer, start by directly generating either 1 or 0, \\then follow with reasons.\\

Here are some examples:\\
C: However, the war times have passed and there are fewer who remember or have lived through those conditions and the \\hardships of life. Now with some people having more money than they actually need there is no strong need to help each \\other out.\\
S: Most of us also live in small apartments, where only the father, mother and the child rent.\\
Answer: 1\\
Concise explanation: "us" in the sentence does not connect semantically with "people" in the context. Hence, "us" should be \\changed to "people".\\

... (14 more examples)\\

C: As there is a saying that try and try until you succeed and success is the stepping stone this is said because by trying new \\things only the man can prove himself to be the successful person he must also be confident of what he is doing.\\
S: If we do the things as we know how to do, it will not cost anything that we cannot gain knowledge of by doing them.\\
Answer: 0\\
Concise explanation: This sentence S is tangential and unnecessary.\\

Now, please generate:\\\\
C: [context]\\
S: [sentence]\\
Answer:
\end{tabular} \\ \hline
\end{tabular}
\end{adjustbox}
\caption{16-shot prompt for GPT-4 cohesion reasoning.}
\label{tab:gpt_cohesion_prompt_16}
\end{table*}

\subsection{Reasoning - Consistency}

For reasoning about the current sentence's consistency, Table~\ref{tab:gpt_consistency_prompt_zero} shows the zero-shot prompt while Table~\ref{tab:gpt_consistency_prompt_16} shows the 16-shot prompt.

\begin{table*}[]
\centering
\begin{adjustbox}{max width=\textwidth}
\begin{tabular}{l}
\hline
\begin{tabular}[c]{@{}l@{}}
You are about to perform the task of consistency detection for the sentences written by second-language English learners. \\In this task, given a sentence S that is incoherent with a context C, you need to output 1 if S contradicts previously \\presented information in C; otherwise, output 0. Now, please generate the answer:\\\\
C: [context]\\
S: [sentence]\\
Answer:
\end{tabular} \\ \hline
\end{tabular}
\end{adjustbox}
\caption{Zero-shot prompt for GPT-4 consistency reasoning.}
\label{tab:gpt_consistency_prompt_zero}
\end{table*}

\begin{table*}[]
\centering
\begin{adjustbox}{max width=\textwidth}
\begin{tabular}{l}
\hline
\begin{tabular}[c]{@{}l@{}}
You are about to perform the task of consistency detection for the sentences written by second-language English learners. \\In this task, given a sentence S that is incoherent with a context C, you need to output 1 if S contradicts previously \\presented information in C; otherwise, output 0.\\

Here are some examples:\\
C: Then, I wanna indicate that young people can study many things that are interesting or exciting things for young people.\\
S: About students, they can learn various fields that students want to study.\\
Answer: 0\\
Concise explanation: Sentence S does not contradict previously presented information in C. S is incoherent with C because \\"students" does not connect semantically with "young people" in the context.\\ 

... (14 more examples)\\

C: As there is a saying that try and try until you succeed and success is the stepping stone this is said because by trying \\new things only the man can prove himself to be the successful person he must also be confident of what he is doing.\\
S: If we do the things as we know how to do, it will not cost anything that we cannot gain knowledge of by doing them.\\
Answer: 0\\
Concise explanation: Sentence S does not contradict previously presented information in C. This sentence S is tangential \\and unnecessary.\\

Now, please generate:\\\\
C: [context]\\
S: [sentence]\\
Answer:
\end{tabular} \\ \hline
\end{tabular}
\end{adjustbox}
\caption{16-shot prompt for GPT-4 consistency reasoning.}
\label{tab:gpt_consistency_prompt_16}
\end{table*}

\subsection{Reasoning - Relevance}

For reasoning about the current sentence's relevance, Table~\ref{tab:gpt_relevance_prompt_zero} shows the zero-shot prompt while Table~\ref{tab:gpt_relevance_prompt_16} shows the 16-shot prompt.

\begin{table*}[]
\centering
\begin{adjustbox}{max width=\textwidth}
\begin{tabular}{l}
\hline
\begin{tabular}[c]{@{}l@{}}
In this task, given a sentence S that is incoherent with a context C, you need to output 1 if S is incoherent with C \\because of a lack of relevance based on the following instructions; otherwise, output 0. You should output 1 only if: \\a) sentence S introduces information that is completely irrelevant to context C, or \\b) sentence S introduces information that is either tangential or slightly irrelevant to context C.\\\\
C: [context]\\
S: [sentence]\\
Answer:
\end{tabular} \\ \hline
\end{tabular}
\end{adjustbox}
\caption{Zero-shot prompt for GPT-4 relevance reasoning.}
\label{tab:gpt_relevance_prompt_zero}
\end{table*}

\begin{table*}[]
\centering
\begin{adjustbox}{max width=\textwidth}
\begin{tabular}{l}
\hline
\begin{tabular}[c]{@{}l@{}}
In this task, given a sentence S that is incoherent with a context C, you need to output 1 if S is incoherent with C \\because of a lack of relevance based on the following instructions; otherwise, output 0. You should output 1 only if: \\a) sentence S introduces information that is completely irrelevant to context C, or \\b) sentence S introduces information that is either tangential or slightly irrelevant to context C.\\

Here are some examples:\\
C: Then, I wanna indicate that young people can study many things that are interesting or exciting things for young \\people.\\
S: About students, they can learn various fields that students want to study.\\
Answer: 0\\
Concise explanation: Sentence S introduces information that is relevant to context C. S is incoherent with C \\because "students" does not connect semantically with "young people" in the context. \\

... (14 more examples)\\

C: As there is a saying that try and try until you succeed and success is the stepping stone this is said because by \\trying new things only the man can prove himself to be the successful person he must also be confident of what he \\is doing.\\
S: If we do the things as we know how to do, it will not cost anything that we cannot gain knowledge of by doing \\them.\\
Answer: 1\\
Concise explanation: Sentence S introduces information that is irrelevant and tangential to context C. This sentence \\S is tangential and unnecessary. Hence, Sentence S is incoherent with context C because of the relevance issue.\\

Now, please generate:\\\\
C: [context]\\
S: [sentence]\\
Answer:
\end{tabular} \\ \hline
\end{tabular}
\end{adjustbox}
\caption{16-shot prompt for GPT-4 relevance reasoning.}
\label{tab:gpt_relevance_prompt_16}
\end{table*}

\subsection{Reasoning - Others}

For reasoning about the current sentence's incoherence that belongs to neither of the above three categories, and instead is a disagreement of the sentence topic with its comment, Table~\ref{tab:gpt_others_prompt_zero} shows the zero-shot prompt while Table~\ref{tab:gpt_others_prompt_16} shows the 16-shot prompt.

\begin{table*}[]
\centering
\begin{adjustbox}{max width=\textwidth}
\begin{tabular}{l}
\hline
\begin{tabular}[c]{@{}l@{}}
In this task, given a sentence S that is incoherent with a context C, you need to output 1 if S is incoherent with C \\because of the disagreement between the topic and the comment of sentence S; otherwise, output 0. Specifically, \\you should output 1 only if the comment of sentence S does not agree with the topic of the sentence itself. Now, \\please generate the answer:\\\\
C: [context]\\
S: [sentence]\\
Answer:
\end{tabular} \\ \hline
\end{tabular}
\end{adjustbox}
\caption{Zero-shot prompt for GPT-4 reasoning for other categories, e.g. topic-comment disagreement.}
\label{tab:gpt_others_prompt_zero}
\end{table*}

\begin{table*}[]
\centering
\begin{adjustbox}{max width=\textwidth}
\begin{tabular}{l}
\hline
\begin{tabular}[c]{@{}l@{}}
In this task, given a sentence S that is incoherent with a context C, you need to output 1 if S is incoherent with C \\because of the disagreement between the topic and the comment of sentence S; otherwise, output 0. Specifically, \\you should output 1 only if the comment of sentence S does not agree with the topic of the sentence itself.\\

Here are some examples:\\
C: I, however, think in terms of physical and mental factors young people are superior to older people.\\
S: For example, in the case of sports young people can run and jump, and they can train their muscles that are \\used in each sport such as transitional sports or silence sports.\\
Answer: 1\\
Concise explanation: The comment of sentence S does not agree with the topic of the sentence. "transitional \\sports" and "silence sports" are not consistent with each other in sentence S itself, and they also do not agree \\with the topic of the sentence. Hence, Sentence S is incoherent with context C because of the disagreement \\between the topic and the comment of sentence S.\\

... (14 more examples)\\

C: As there is a saying that try and try until you succeed and success is the stepping stone this is said because by \\trying new things only the man can prove himself to be the successful person he must also be confident of what \\he is doing.\\
S: If we do the things as we know how to do, it will not cost anything that we cannot gain knowledge of by doing \\them.\\
Answer: 0\\
Concise explanation: The comment of sentence S agrees with the topic of the sentence. This sentence S is \\tangential and unnecessary. Hence, Sentence S is incoherent with context C because of the relevance issue. \\

Now, please generate:\\\\
C: [context]\\
S: [sentence]\\
Answer:
\end{tabular} \\ \hline
\end{tabular}
\end{adjustbox}
\caption{16-shot prompt for GPT-4 reasoning for other categories, e.g. topic-comment disagreement.}
\label{tab:gpt_others_prompt_16}
\end{table*}

\section{GPT-4 Judge Prompt}
\label{sec:gpt_judge_prompt}
% Note that we suggest "DELETE" if sentence S is tangential because having sentence S following context C is unnecessary.

For the incoherence rewriting task, we employ GPT-4 as a judge to conduct the pairwise evaluations and determine which one is better than the other, as explained in Section~\ref{sec:incoherence_rewriting}. The prompt we use is shown in Table~\ref{tab:gpt_prompt_judge}.

\begin{table*}[]
\centering
\begin{adjustbox}{max width=\textwidth}
\begin{tabular}{l}
\hline
\begin{tabular}[c]{@{}l@{}}
You are an English teacher aiming to improve coherence in student writing. \\You need to evaluate and select the best system based on the coherence of their outputs to a given instruction. \\This process will be used to create a leaderboard reflecting the most accurate and human-preferred answers.\\

I require a leaderboard for various systems. I'll provide you with prompts given to these models and their corresponding outputs. \\Your task is to assess these responses, and select the model that produces the most coherent output from a English teacher's perspective. \\\\

\#\# Instruction \\\\ You are about to perform the task of sentence rewriting for the sentences written by second-language English learners. \\In this task, given a context C and a sentence S, where S is incoherent with C, you need to rewrite sentence S to make it coherent with C \\according to the following instructions: The rewrite should be as minimal as possible. [reason\_texts]. \\C: [context] \\S: [sentence] \\Answer:\\\\\#\# Model Outputs\\\\Here are the unordered outputs from the models. Each output is associated with a specific model, identified by a unique model identifier.\\\\

\#\#\# model\_identifier: "m" \\\#\#\#\# output: [output\_1] \\\\\#\#\# model\_identifier: "M" \\\#\#\#\# output: [output\_2] \\\\\#\# Task \\\\Evaluate the models based on the coherence of their outputs to the given context C, and select the model that generated the most coherent \\output. \\The output from the model should rewrite the incoherent sentence S as minimally as possible. \\Retaining awkward phrasing or minor grammar errors from sentence S is acceptable as long as the output is coherent with context C. \\Answer by first providing a concise explanation and then end your answer by providing the model identifier of the most coherent output. \\We will use the last character of your output `output[-1]' as the name of the best model, so make sure you finish with the token of the \\model identifiers and nothing else: `m' or `M' (no quotes, no dots, no backticks, no new lines, ...). For example: \\\\\#\#\# Concise explanation \\...some text... \\\\\#\#\# Which is best, m or M? \\1\\\\Now is your turn. \\\\\#\# Your answer: "Concise explanation" followed by "Which is best, m or M?"
\end{tabular} \\ \hline
\end{tabular}
\end{adjustbox}
\caption{The prompts for GPT-4 as a judge to conduct the pairwise comparisons between model outputs.}
\label{tab:gpt_prompt_judge}
\end{table*}

\end{document}